\documentclass[journal]{IEEEtran}
\usepackage{times}
\usepackage{amsmath}
\usepackage[pdftex]{graphicx}
\usepackage{subfigure}
\usepackage{amsmath,amssymb,amsopn,amstext,amsfonts}
\usepackage[nolist,nohyperlinks]{acronym}
\usepackage{bm}
\usepackage{cancel}
\usepackage[space]{cite}
\usepackage{pdfsync}
\usepackage{balance}
\usepackage{color}
\usepackage{algorithm}
\usepackage{algorithmic}
\usepackage{url}
\usepackage{booktabs}
\usepackage{threeparttable}
\usepackage[linkcolor=black,citecolor=black,urlcolor=black,colorlinks=true]{hyperref}
\usepackage{multirow}
\usepackage{graphicx}
\IEEEoverridecommandlockouts

\newcommand\inv[1]{#1\raisebox{1.15ex}{$\scriptscriptstyle-\!1$}}
\DeclareMathOperator*{\argmaxA}{arg\,max} % Jan Hlavacek
\DeclareMathOperator*{\argminA}{arg\,min}

\graphicspath{{./figures/}}
\DeclareGraphicsExtensions{.pdf,.png,.jpg,}

\begin{acronym}
	\acro{SfM}{Structure from Motion}
	\acro{TCXO}{Temperature Controlled X'tal(crystal) Oscillator}
	\acro{GPST}{GPS Time}
	\acro{GNSS}{Global Navigation Satellite System}
	\acro{AR}{Augmented Reality}
	\acro{VIN}{Visual-Inertial Navigation}
	\acro{VIO}{Visual-Inertial Odometry}
	\acro{UAV}{Unmanned Aerial Vehicle}
	\acro{IMU}{Inertial Measurement Unit}
	\acro{DoF}{Degree of Freedom}
	\acro{ECI}{Earth-Centered Inertial}
	\acro{ECEF}{Earth-Centered, Earth-Fixed}
	\acro{ENU}{East-North-Up}
	\acro{EKF}{Extended Kalman Filter}
	\acro{UKF}{Unscented Kalman Filter}
	\acro{ToF}{Time of Flight}
	\acro{PRN}{Pseudo Random Noise}
	\acro{SPP}{Single Point Positioning}
	\acro{RF}{Radio Frequency}
	\acro{EMI}{ElectroMagnetic Interference}
	\acro{RTK}{Real-Time Kinematic positioning}
	\acro{RMSE}{Root Mean Square Error}
	\acro{PPP}{Precise Point Positioning}
	\acro{PPS}{Pluse Per Second}
\end{acronym}

\begin{document}
\title{GVINS: Tightly Coupled GNSS-Visual-Inertial Fusion for Smooth and Consistent State Estimation}

\author{Shaozu~Cao, %~\IEEEmembership{Member,~IEEE,}
        Xiuyuan~Lu, %~\IEEEmembership{Member,~IEEE,}
        and~Shaojie~Shen%~\IEEEmembership{Member,~IEEE,}% <-this % stops a space
\thanks{All authors are with the Department of Electronic and Computer Engineering,
Hong Kong University of Science and Technology, Hong Kong, China.
{\tt\small \{scaoad, xluaj\}@connect.ust.hk, eeshaojie@ust.hk}. \textit{(Corresponding author: Shaozu Cao.)}}% <-this % stops a space
% \thanks{Manuscript received April 19, 2005; revised August 26, 2015.}
}

\maketitle

\begin{abstract}
Visual-Inertial odometry (VIO) is known to suffer from drifting especially over long-term runs. In this paper, we present GVINS, a non-linear optimization based system that tightly fuses GNSS raw measurements with visual and inertial information for real-time and drift-free state estimation. Our system aims to provide accurate global 6-DoF estimation under complex indoor-outdoor environment where GNSS signals may be intermittent or even totally unavailable. To connect global measurements with local states, a coarse-to-fine initialization procedure is proposed to efficiently calibrate the transformation online and initialize GNSS states from only a short window of measurements. The GNSS code pseudorange and Doppler shift measurements, along with visual and inertial information, are then modelled and used to constrain the system states in a factor graph framework. For complex and GNSS-unfriendly areas, the degenerate cases are discussed and carefully handled to ensure robustness. Thanks to the tightly-coupled multi-sensor approach and system design, our system fully exploits the merits of three types of sensors and is capable to seamlessly cope with the transition between indoor and outdoor environments, where satellites are lost and reacquired. We extensively evaluate the proposed system by both simulation and real-world experiments, and the result demonstrates that our system substantially eliminates the drift of VIO and preserves the local accuracy in spite of noisy GNSS measurements. The challenging indoor-outdoor and urban driving experiments verify the availability and robustness of GVINS in complex environments. In addition, experiments also show that our system can gain from even a single satellite while conventional GNSS algorithms need four at least.

\end{abstract}


\section{Introduction}

\IEEEPARstart{L}{ocalization} is an essential functionality for many spatial-aware applications, such as autonomous driving, \ac{UAV} navigation and \ac{AR}. Estimating system states with various sensors has been widely studied for decades. Among these, the sensor fusion approach has been more and more popular in recent years. Due to the complementary properties provided by heterogeneous sensors, sensor fusion algorithms can significantly improve the accuracy and robustness of the state estimation system.

The camera provides rich visual information with only a low cost and small footprint, thus attracting much attention from both computer vision and robotics area. Combined with a MEMS IMU, which offers high frequency and outlier-free inertial measurement, \ac{VIN} algorithms can often achieve high accuracy and be more robust in complex environments. Nevertheless, both camera and IMU operate in the local frame and it has been proven that the \ac{VIN} system has four unobservable directions\cite{huang2014towards}, namely $x$, $y$, $z$ and $yaw$. Thus the odometry drift is inevitable for any VIN system. On the other hand, \ac{GNSS} provides a drift-free and global-aware solution for localization tasks, and has been extensively used in various scenarios. \ac{GNSS} signal is freely available and conveys the range information between the receiver and satellites. With at least 4 satellites being tracked simultaneously, the receiver is able to obtain its unique coordinate in the global Earth frame. Considering the complementary characteristics between the \ac{VIN} and \ac{GNSS} system, it seems natural that improvements can be made by fusing information from both systems together.

However, many challenges exist during the fusion of two systems. Firstly, a stable initialization from noisy GNSS measurement is indispensable. Among quantities need to be initialized, the 4-DoF transformation between the local \ac{VIN} frame and the global GNSS frame is essential. The transformation is necessary to associate measurements from local and global systems together. Unlike the extrinsic transformation between camera and IMU, this transformation cannot be offline calibrated because each time the \ac{VIN} system starts such transformation will vary. In addition, one-shot alignment using a portion of sequence does not work well as the drift of the fusion system makes such alignment invalid during GNSS outage situations. Thus, an online initialization and calibration between local frame and global frame is necessary to fuse heterogeneous measurements and cope with complex indoor-outdoor environments. Secondly, the precision of the GNSS measurement does not match with that of the \ac{VIN} system, and various error sources exist during the GNSS signal propagation. In practice, the code pseudorange measurement, which is used for global localization, can only achieve meter-level precision while the \ac{VIN} system is capable to provide centimeter-level estimation over a short range. As a result, the fusion system will be susceptible to the noisy GNSS measurement if not formulated carefully. Thirdly, degeneration happens when the fusion system experiences certain movements such as pure rotation or the number of locked satellites is insufficient. Normally the GNSS-visual-inertial fusion system can offer a drift-free 6-DoF global estimation, but the conclusion no longer holds in degenerate cases. In addition, the transition between indoor and outdoor environments, during which all satellites are lost and gradually reacquired again, also poses challenges to the system design. 

\begin{figure}
    \centering
    \vspace{-1.61cm}
\end{figure}

\begin{figure*}
    \centering
    \includegraphics[width=2.0\columnwidth]{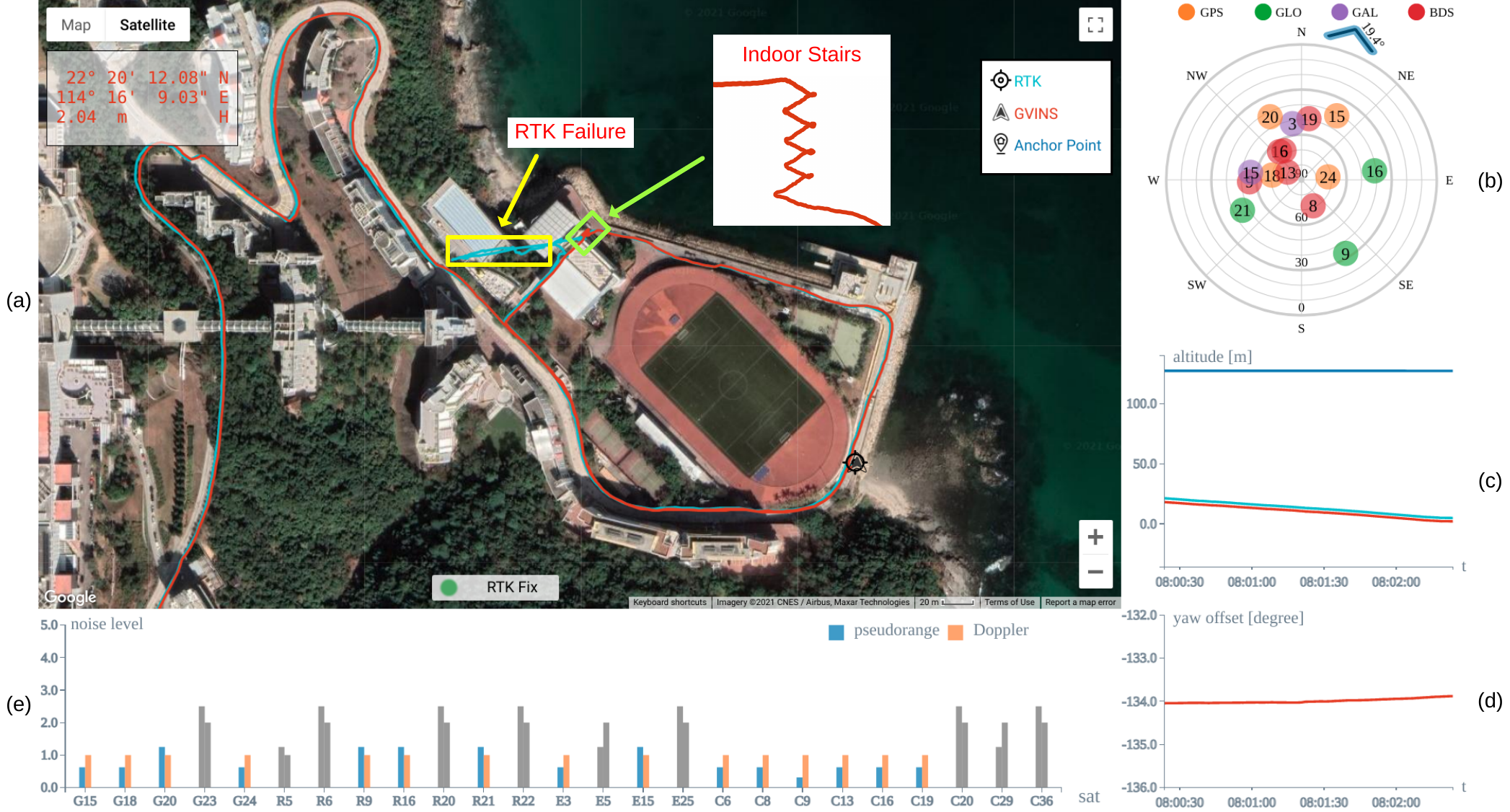}
    \caption{
		\label{fig:teaser} 
		A snapshot of our system in a complex indoor-outdoor environment. The global estimation result is plotted on Google Maps directly and aligns well with the ground truth RTK trajectory as shown in part (a). Part (b) depicts the distribution of satellites with tangential direction representing the azimuth and radial direction being the elevation angle. The blue arrow is a compass-like application which indicts the global yaw orientation of the camera. Subplot (c) and (d) illustrate the altitude information and the local-ENU yaw offset respectively. The measurement noise level of each tracked satellite is shown in part (e). Note that there is an obvious failure on the RTK trajectory when we walk the indoor stairs, while our system can still perform global estimation even in indoor environment. }
\end{figure*}

To address the above-mentioned issues, we propose a non-linear optimization-based system to tightly fuse \ac{GNSS} raw measurements (code pseudorange and Doppler frequency shift) with visual and inertial data for accurate and drift-free state estimation. The 4-DoF transformation between local and global frames is recovered via a coarse-to-fine approach during initialization phase and is further optimized subsequently. To incorporate noisy \ac{GNSS} raw measurements, all \ac{GNSS} constraints are formulated under a probabilistic factor graph in which all states are jointly optimized. In addition, degenerate cases are discussed and carefully handled to ensure robustness. Thanks to the tightly-coupled approach and system design, our system fully exploits the complementary properties among \ac{GNSS}, visual and inertial measurements and is able to provide locally smooth and globally consistent estimation even in complex environments, as shown in Fig.~\ref{fig:teaser}. We highlight the contributions of this paper as follows:

\begin{itemize}
	\item an online coarse-to-fine approach to initialize GNSS-visual-inertial states.
	\item an optimization-based, tightly-coupled approach to fuse visual-inertial data with multi-constellation \ac{GNSS} raw measurements under the probabilistic framework.
	\item a real-time estimator which is capable to provide drift-free 6-DoF global estimation in complex environment where \ac{GNSS} signals may be largely intercepted or even totally unavailable.
	\item an evaluation of the proposed system in both simulation and real-world environments.
\end{itemize}  
For the benefit of the research community, the proposed system\footnote{https://github.com/HKUST-Aerial-Robotics/GVINS}, along with the well-synchronized datasets \footnote{https://github.com/HKUST-Aerial-Robotics/GVINS-Dataset},  have been open-sourced. 

The rest of this work is structured as follows: in Section \ref{sec:literature} we discuss the existing relevant literature. Section \ref{sec:notation} describes the notation and coordinate system involved in the system. In Section \ref{sec:gnss_fundamentals} we briefly introduce relevant background knowledge of \ac{GNSS}. Section \ref{sec:system_overview} shows the structure and workflow of the proposed system. The problem formulation and methodology are illustrated in Section \ref{sec:formulation}. In Section \ref{sec:initialization_and_degeneration} we address the \ac{GNSS} initialization issues and discuss several degenerate cases that degrade the performance of our system. The experiment setup and evaluation are given in Section \ref{sec:experiments}. Finally Section \ref{sec:conclusion} concludes this paper.

\section{Related Work}
\label{sec:literature}
State estimation via multiple sensors fusion approach has been proven to be effective and robust, and there is extensive literature on this area. Among those, we are particularly interested in the combination of small size and low cost sensors such as camera, IMU and \ac{GNSS} receiver, to produce a real-time accurate estimation in the unknown environment.

The fusion of visual and inertial measurement in a tightly-coupled manner can be classified into either filter-based method or optimization-based method. MSCKF\cite{MouRou0704} is an excellent filter-based state estimator which utilizes the geometric constraints between multiple camera poses to efficiently optimize the system states. Based on MSCKF, \cite{LiMou1305} makes improvements on its accuracy and consistency, and \cite{wu2015square} aims to overcome its numerical stability issue especially on mobile devices. Compared with the filter based approach, nonlinear batch optimization method can achieve better performance by re-linearization at the expense of computational cost. OKVIS\cite{LeuFurRab1306} utilizes keyframe-based sliding window optimization approach for state estimation. VINS-Mono\cite{qin2018vins} also optimizes system states within the sliding window but is more complete with online relocalization and pose graph optimization. Since camera and IMU only impose local relative constraint among states, accumulated drift is a critical issue in the VIN system, especially over long-term operation.

As \ac{GNSS} provides absolute measurement in the global Earth frame, incorporating \ac{GNSS} information is a natural way to reduce accumulated drift. In terms of loosely-coupled manner, \cite{6289875}\cite{lynen2013robust} describe state estimation systems which fuse \ac{GNSS} solution with visual and inertial data under the EKF framework. \cite{SheMulMic1405} proposes a UKF algorithm that fuses visual, inertial, LiDAR and \ac{GNSS} solution to produce a smooth and consistent trajectory in different environments. \cite{mascaro2018gomsf}, \cite{8968519} and our previous work VINS-Fusion\cite{qin2019general} fuse the result from local VIO with \ac{GNSS} solution under the optimization framework. In \cite{li2021semi} the authors combine the result from \ac{PPP} with a stereo VIO to achieve low drift estimation. Both the GNSS code and phase measurements are used in their formulation and precise satellite products are utilized to improve the accuracy. All aforementioned works rely on the \ac{GNSS} solution to perform estimation so system failure will occur once the \ac{GNSS} solution is highly corrupted or unavailable in the situation where the number of tracked satellites is below than 4. 

In the line of tightly-coupled GNSS-visual approaches, \cite{gakne2018tightly} tightly fuses \ac{GNSS} code pseudorange data and visual measurements from a sky-pointing camera in the EKF manner. The image from the upward-facing camera is segmented as the sky and non-sky areas. The non-sky areas are used for feature detection and matching. In addition, only GNSS signals coming from sky directions are used to avoid potential multipath effect. However, the upward-facing camera means that their system cannot work in an open-sky scenario and is only suitable for urban environments. In addition, the transformation between the local vehicle frame and the global frame is assumed known in their work. In \cite{schreiber2016vehicle} the authors proposed a system that tightly combine the stereo visual odometry with the GNSS code pseudorange and Doppler shift measurements using the EKF framework. Three driving tests with moderate distance were conducted to evaluate their system. However, only horizontal errors are reported in their first data sequence and the majority of their experiments are just qualitatively analysed. 

There are also some works on tightly fusing \ac{GNSS} raw measurement with visual and inertial information. \cite{6851506}, \cite{li2019tight} and \cite{yoder2020multi} combine camera, IMU and \ac{GNSS} RTK measurements under the EKF framework for localization. The RTK solution, which usually owns centimeter-level accuracy, requires a static \ac{GNSS} reference station with known position as infrastructure. \cite{5507322} and \cite{won2014gnss} investigate the performance of the fusion system in cluttered urban environment where less than 4 satellites are tracked. However, the transformation between local and global frames is not handled and the scale of their real-world experiments is limited. In addition, the result of the underlying VIN system in \cite{won2014gnss}, as is tested in standalone mode, shows large drift over a short period of time. Recently we found a similar work\cite{DBLP:journals/corr/abs-2010-11675} that tightly fuse \ac{GNSS} raw measurements with visual-inertial SLAM. An RMSE error of $14.33 ~m$ is reported on the longest sequence ($5.9 ~km$) in their evaluation, whereas the value is only $4.51 ~m$ for our system, even on a more challenging urban driving sequence with a total distance of $22.9 ~km$. In GNSS-unfriendly areas where the number of GNSS measurements becomes insufficient, \cite{DBLP:journals/corr/abs-2010-11675} drops all GNSS measurements which may still benefit the estimator as shown in our experiments. In addition, the indoor environments within the sequence such as tunnels cannot be handled by their system, which again limits the potential of the tightly multi-sensor fusion approach.

To this end, we aim to build a robust and accurate state estimator with \ac{GNSS} raw measurements, visual and inertial data tightly fused. By leveraging the global measurement from \ac{GNSS}, the accumulated error from visual-inertial system will be eliminated. The transformation between the local and global frame will be estimated without any offline calibration. The system is capable to work in complex indoor and outdoor environments and achieves local smoothness and global consistency.

\begin{figure}
    \centering
    \includegraphics[width=0.6\columnwidth]{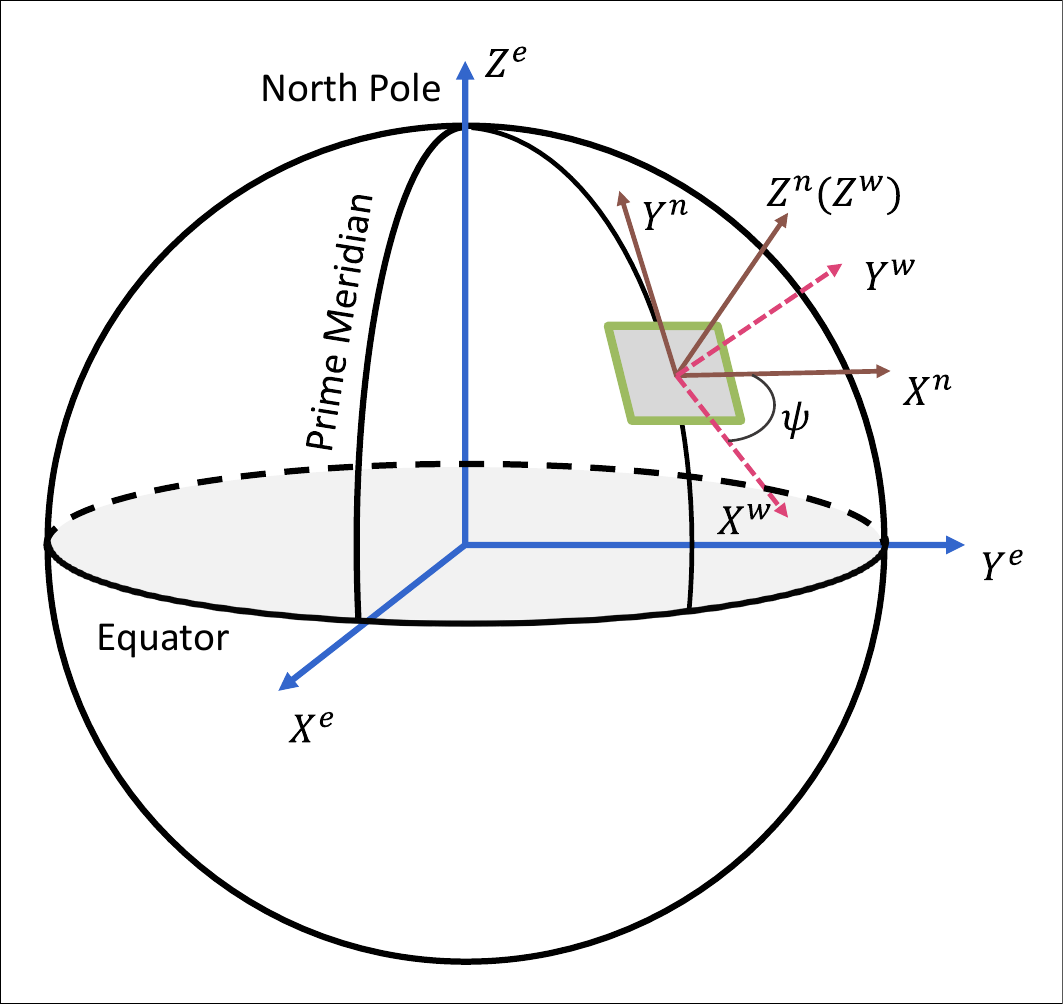}
    \caption{
        \label{fig:coordinate_systems} 
		An illustration of the local world, ECEF and ENU frames.}
	\vspace{-0.3cm}
\end{figure}

\section{Notation and Definitions}
\label{sec:notation}

\subsection{Frames}
The spatial frames involved in our system consist of:

\begin{itemize}
\item Sensor Frame: 
The sensor frame is attached to the sensor and is a local frame in which sensor reports its reading. In our system, sensor frames include the camera frame $(\cdot)^c$ and the IMU frame $(\cdot)^i$, and we choose IMU frame as our estimation target frame and denote it as body frame $(\cdot)^b$.

\item Local World Frame: 
We represent the conventional frame in which visual-inertial system operates as the local world frame $(\cdot)^w$. In VIN system, the origin of the local world frame is arbitrarily set and the z axis is often chosen to be gravity-aligned
as illustrated in Fig.~\ref{fig:coordinate_systems}.

\item ECEF Frame: 
The \ac{ECEF} frame $(\cdot)^e$ is a Cartesian coordinate system that is fixed with respect to Earth. As shown in Fig.~\ref{fig:coordinate_systems}, the origin of ECEF frame is attached to the center of mass of Earth. The x-y plane coincides with the Earth's equatorial plane with x-axis pointing to the prime meridian. The z-axis is chosen to be perpendicular to the Earth's equatorial plane in the direction of the geographical North Pole. Finally the y-axis is taken to make ECEF frame a right-handed coordinate system. In this paper we use the WGS84 realization of ECEF frame. 

\item ENU Frame: 
In order to connect the local world and global ECEF frames, a semi-global frame, ENU, is introduced. The x, y, z axis of the ENU frame $(\cdot)^n$ point to the east, north, and up direction respectively (Fig.~\ref{fig:coordinate_systems}). Given a point in ECEF frame, a unique ENU frame can be determined with its origin sitting on that point. Note that the z axis of ENU frame is also gravity aligned.

\item ECI Frame: 
The \ac{ECI} frame is an inertial coordinate system with the center of mass of the Earth as its origin. The three axes of the ECI frame $(\cdot)^E$ are taken to point in fixed directions with respect to the stars, i.e., do not rotate with the Earth. The GNSS signal travels in straight line in the ECI frame, which can greatly simplify the formulation. In this paper the ECI frame is formed by freezing the ECEF frame at the time of reception of the GNSS signal.

\end{itemize}

In terms of temporal frames, GNSS data is tagged in GNSS time system (for example, GPS time), while visual and inertial measurements are marked in the local time system. We assume that these two time systems are aligned beforehand and do not distinguish them accordingly.

\subsection{Notation}
In this paper we use $\mathbf{R}^z_a$ and $\mathbf{p}^z_a$ to denote the rotational and translational part of the transformation from frame $a$ to frame $z$. For rotational part, the corresponding Hamilton quaternion $\mathbf{q}^z_a$ is also used, with $\otimes$ representing its multiplication operation. We use subscript to refer a moving frame at a specific time instance. For example, $\mathbf{R}^z_{a_t}$ stands for the rotation from the moving frame $a$ at time $t$ to the fixed frame $z$. 

For constant quantities, we use $\mathbf{g}^w$ to represent the gravity vector in the local world frame. $c$ is the speed of light in vacuum and $\omega_E$ stands for the angular velocity of the Earth. 
\subsection{States}
The system states to be estimated include:
\begin{itemize}
	\item the position $\mathbf{p}^w_b$ and orientation $\mathbf{q}^w_b$ of the body frame with respect to the local world frame,
	\item the velocity $\mathbf{v}^w_b$, accelerometer bias $\mathbf{b}_a$ and gyroscope bias $\mathbf{b}_w$,
	\item the inverse depth $\rho$ for each feature,
	\item the yaw offset $\psi$ between the local world frame and ENU frame, receiver clock bias $\delta \mathbf{t}$ and receiver clock drifting rate $\delta \dot{t}$. Because our system support all four constellations, the clock biases for GPS, GLONASS, Galileo and BeiDou are estimated separately. Note that the receiver clock drifting rate for each constellation is the same.
\end{itemize} 

Our system adopt a sliding window optimization manner and states $\mathcal{X}$ inside the window can be summarized as

\begin{subequations}
	\label{eq:states}
	\begin{align}
		\mathcal{X}    &= \left [ \mathbf{x}_0,\,\mathbf{x}_{1},\, \cdots \,\mathbf{x}_{n},\,  \rho_0,\,\rho_{1},\, \cdots \,\rho_{m},\, \psi \,\right ] \\
		\mathbf{x}_k   &= \left [ \mathbf{p}^w_{b_{t_k}},\,\mathbf{v}^w_{b_{t_k}},\,\mathbf{q}^w_{b_{t_k}}, \,\mathbf{b}_a, \,\mathbf{b}_w,\,  \delta \mathbf{t}, \,\dot{\delta t} \,\right ],\, k\in [0,n] \\
		\delta \mathbf{t} &= \left [ \delta t_{G},\, \delta t_{R},\, \delta t_{E},\, \delta t_{C} \right ] ~,
	\end{align}
\end{subequations}
where $n$ is the window size and $m$ is the number of feature points in the window. The four components in $\delta \mathbf{t}$ correspond to receiver's clock biases with respect to the time of GPS, GLONASS, Galileo and BeiDou respectively.

\section{GNSS Fundamentals}
\label{sec:gnss_fundamentals}
Since our system requires GNSS raw measurement processing, background knowledge about GNSS is necessary. In this section, we first give an overview about GNSS. Then two types of raw measurements, namely code pseudorange and Doppler shift, are introduced and modelled. Finally the principle of \ac{SPP} algorithm for global localization is described in the end of this section.

\subsection{GNSS Overview}
\label{ssec:gnss_overview}
\begin{figure}
    \centering
    \includegraphics[width=1.0\columnwidth]{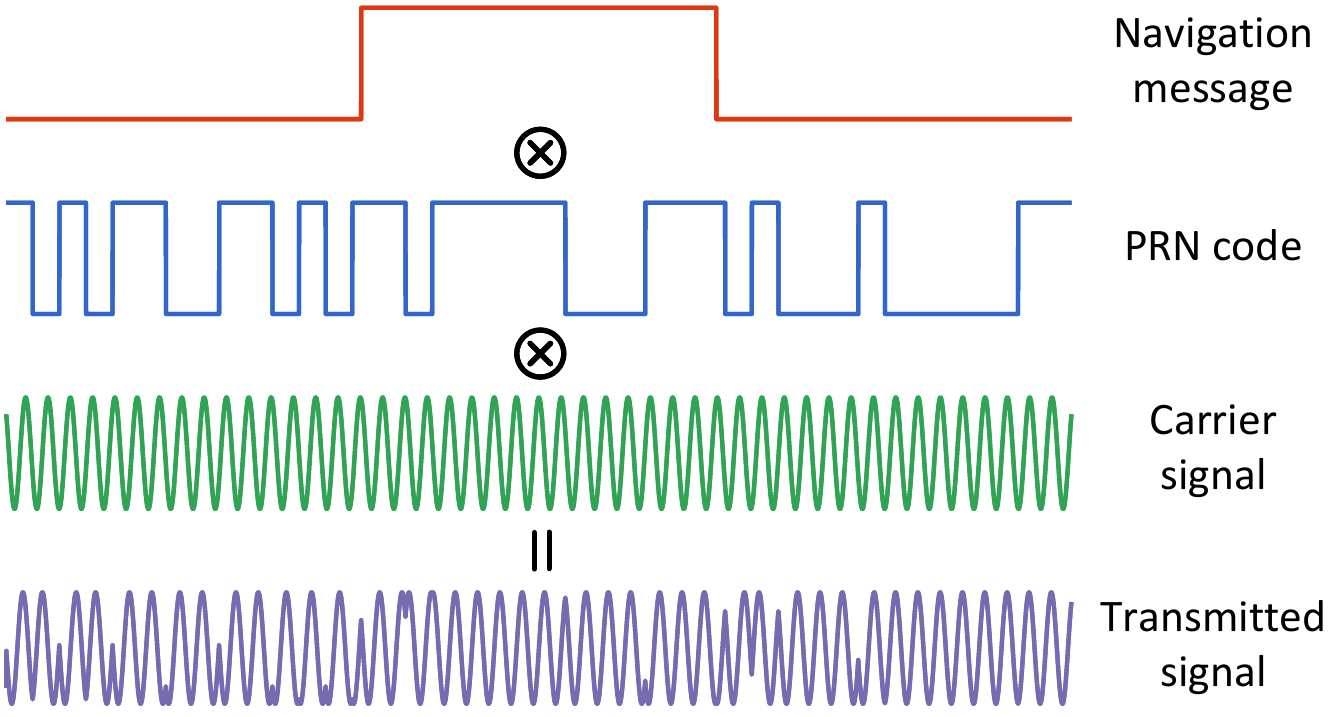}
    \caption{
        \label{fig:gnss_signal} 
		The hierarchical structure of the GPS L1C signal. The navigation message first mixes with the satellite-specific PRN code, and then the resulting sequence is used to modulate the high frequency carrier signal. The final signal is transmitted by the satellite and captured by the receiver, which applies a reverse process to obtain the measurement and retrieve the message.}
% 	\vspace{-0.3cm}
\end{figure}

Global Navigation Satellite System (GNSS), as its name suggests, is a satellite-based system which is capable to provide global localization service. Currently there are four independent and fully operational systems, namely GPS, GLONASS, Galileo and BeiDou. The navigation satellite continuously transmits radio signal from which the receiver can uniquely identify the satellite and retrieve the navigation message. Taking the GPS L1C signal as an example, the final transmitted signal is composed of three layers, as illustrated in Fig.~\ref{fig:gnss_signal}. The navigation message contains parameters of the orbit, corrections of the clock error, coefficients of ionospheric delay and other information related to satellite's status. The orbit parameters, also know as ephemeris, contains 14 variables and is used to calculate the satellite's ECEF coordinate at a particular time. The satellite's clock error is modelled as a second-order polynomial, i.e., with 3 parameters. Each satellite is assigned a unique \ac{PRN} code that repeats every 1 millisecond. The 50 bit/s navigation message is first exclusive-ored with the PRN code and then used to modulate the high frequency carrier signal. After receiving the signal, the receiver obtains the Doppler shift (Section.~\ref{ssec:Doppler_introduction}) by measuring the frequency difference between the received one and designed one. The code pseudorange measurement (Section.~\ref{ssec:pseudo_introduction}) is inferred from the PRN code shift which indicts the propagation time. Finally the navigation message is uncovered by a reverse demodulation process.

\subsection{Code Pseudorange Measurement}
\label{ssec:pseudo_introduction}
Upon the reception of the signal, the \ac{ToF} of the signal is measured from the PRN code shift. By multiplying with the speed of light, the receiver obtains code pseudorange measurement. The code pseudorange is prefixed with ``pseudo" because it not only contains the geometric distance between the satellite and the receiver, but also includes various errors during the signal generation, propagation and processing. 

The error source on the satellite side mainly consists of satellite orbit and clock error. The orbit error comes from the influence of other celestial objects which are not precisely modelled by the ephemeris, and the clock error is the result of imperfect satellite onboard atomic clock with respect to the standard system time. The orbit and clock errors are monitored and constantly corrected by the system control segment. During the signal propagation from satellite to receiver, it goes through the ionosphere and troposphere, where the speed of the electromagnetic signal is no longer as same as that in vacuum and the signal gets delayed according to the atmosphere components and propagation path. The phenomenon that signal reaches the receiver with different ways, known as multipath effect, may occur and add extra delay especially for low elevation satellites. When the signal arrivals, the ToF is calculated by comparing the signal transmission time, which is marked by the satellite's atomic clock, with the receiver's less accurate local clock time. Thus the range information is also offset by the receiver clock error with respect to the GNSS system time. In conclusion, the code pseudorange measurement can be modelled as
\begin{equation}
	\label{eq:code_pseudorange}
	\begin{split}
		\tilde{P}_{r}^{s} = & \| \mathbf{p}^{E}_{s} - \mathbf{p}^{E}_{r}\| + c \left ( \bm{\zeta}_s^T \delta \mathbf{t} - \Delta t^{s} \right ) \\ & + T_{r}^{s} + I_{r}^{s} + M_{r}^{s} + \epsilon_{r}^{s} ~,
	\end{split}
\end{equation}
where $\mathbf{p}^{E}_{s}$ and $\mathbf{p}^{E}_{r}$ is the ECI coordinate of the satellite $s$ and receiver $r$, respectively. $\bm{\zeta}_s$ is designed to be a $4 \times 1$ indictor vector with the corresponding satellite constellation entity being $1$ and other three entities being $0$. $\Delta t^{s}$ is the satellite clock error, which can be calculated from the broadcast navigation message. $T_{r}^{s}$ and $I_{r}^{s}$ stand for the tropospheric and the ionospheric delay respectively. We use $M_{r}^{s}$ to denote the delay caused by multipath effect and $\epsilon_{r}^{s}$ for the measurement noise. Here the delay terms $T_{r}^{s}$, $I_{r}^{s}$ and $M_{r}^{s}$ are expressed in unit of length, i.e., multiplied by $c$.

\subsection{Doppler Measurement}
\label{ssec:Doppler_introduction}
The Doppler frequency shift is measured from the difference between the received carrier signal and the designed one, and it reflects the receiver-satellite relative motion along the signal propagation path. Due to the characteristic of the GNSS signal structure, the accuracy of the Doppler measurement is usually an order of magnitude higher than that of code pseudorange. The Doppler shift is modelled as:

\begin{equation}
	\label{eq:Doppler}
	\begin{split}
		\Delta \tilde{f}^{s}_{r} &= -\frac{1}{\lambda} \left[ {\bm{\kappa}_{r}^{s}}^{T} (\mathbf{v}^{E}_{s} - \mathbf{v}^{E}_{r}) + c (\dot{\delta t} - \dot{\Delta t^{s}}) \right] + \eta_{r}^{s} ~,
	\end{split}
\end{equation}
where $\mathbf{v}^{E}_{r}$ and $\mathbf{v}^{E}_{s}$ represent the receiver's and satellite's velocity in ECI frame respectively. We use $\lambda$ to denote the wavelength of the carrier signal, and $\bm{\kappa}_{r}^{s}$ for the unit vector from receiver to satellite in ECI frame. $\dot{\Delta t^{s}}$ is the drift rate of the satellite clock error which is reported in the navigation message, and finally $\eta_{r}^{s}$ represents the Doppler measurement noise.

\subsection{SPP Algorithm}
\label{ssec:spp_algorithm}
The Single Point Positioning (SPP) algorithm utilizes code pseudorange measurements to determine the 3-DOF global position of the GNSS receiver via trilateration. Thus in theory the coordinate of the receiver can be obtained by the aid of 3 different satellites. However, as mentioned in Section \ref{ssec:pseudo_introduction}, code pseudorange measurement is offset by the receiver clock bias. Because the receiver clock bias can cause an error of hundreds of kilometers, it must be estimated along with the location in order to get a reasonable result. To this end, at least 4 code pseudorange measurements are required to fully constrain the 3-DOF global position and receiver clock bias. Because different navigation systems use different time references, there exists clock offset between different systems. Additional measurements are necessary in order to estimate the inter-system clock offset if the satellites are from multiple constellations. To summarize, at least $(N+3)$ satellites are required to be simultaneously tracked in order to obtain the uniquely localize the receiver, where $N$ is the number of constellations among the tracked satellites. 

\begin{figure}
    \centering
	\includegraphics[width=1.0\columnwidth]{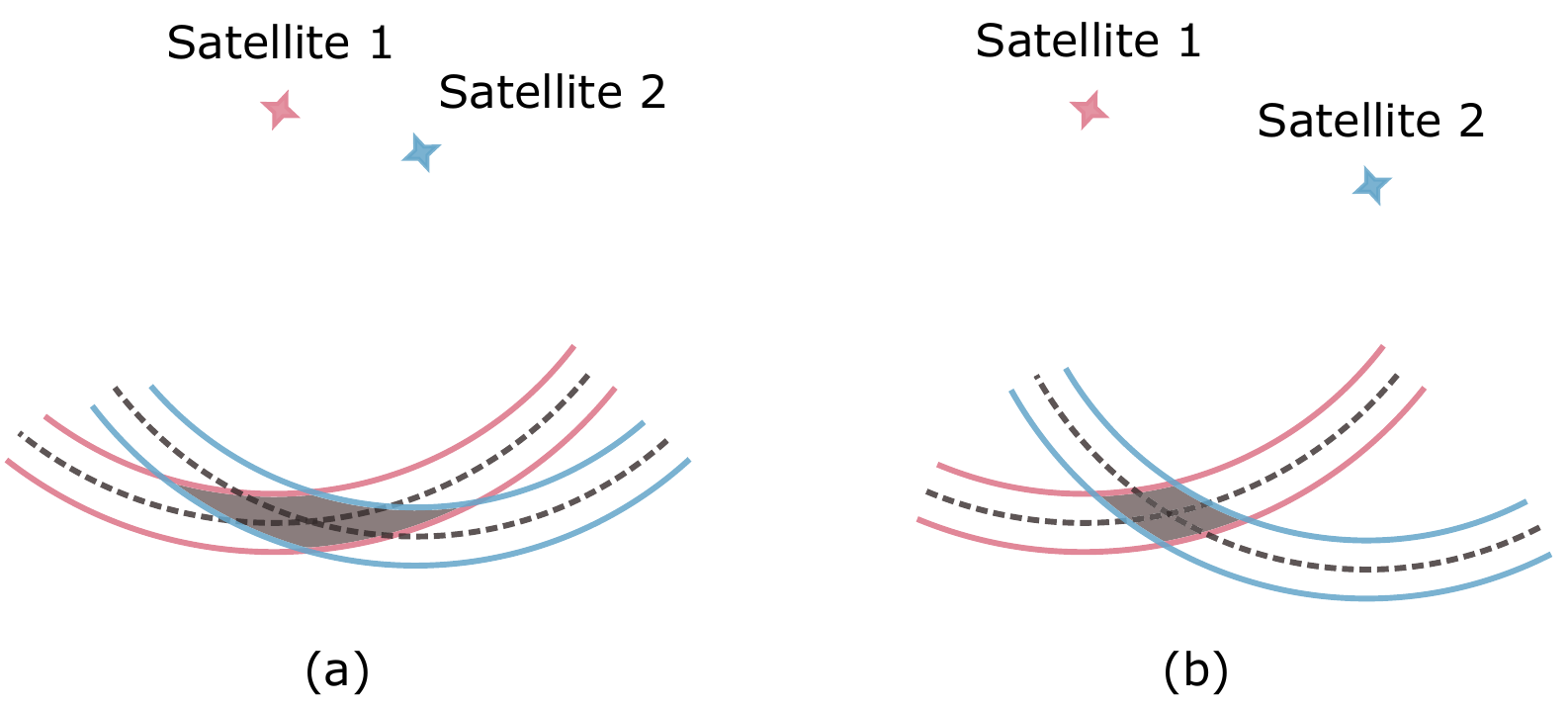}
    \caption{
        \label{fig:sat_distribution} 
        A simplified 2D illustration of how satellites distribution affects the uncertainty of SPP solution. Here we assume the time between the receiver and satellite are synchronized thus two satellites are enough for localization. The dash line represents ground truth range while the area in between the two solid lines denotes the possible noisy measurement. The uncertainties of SPP solutions are represented by the shadows.}
% 	\vspace{-0.5cm}
\end{figure}

After collecting enough measurements, constraints from Eq.~\ref{eq:code_pseudorange} are stacked together to form a series of equations with $\mathbf{p}^{E}_{r}$ and $\delta \mathbf{t}$ unknown. Corrections are applied to code pseudorange measurement making it only a function of $\mathbf{p}^{E}_{r}$ and $\delta \mathbf{t}$. In our system the tropospheric delay $T_{r}^{s}$ is estimated by Saastamoinen model\cite{saastamoinen1972contributions}, and ionospheric delay $I_{r}^{s}$ is computed using Klobuchar model\cite{klobuchar1987ionospheric} and parameters in the ephemeris. By excluding the low elevation satellites, we ignore the delay $M_{r}^{s}$ caused by multipath effect. In practice, more than $(N+3)$ measurements will be used and the solution is obtained by minimizing the sum of the squared residuals. As is shown in \cite{kaplan2005understanding}, the noise of the SPP solution not only depends on the measurement noise but also has a relationship with the geometric distribution of satellites. A simplified 2D case in Fig.~\ref{fig:sat_distribution} shows the effect of satellites distribution on the noise characteristic of the final solution. Thus the performance of SPP algorithm will be better with evenly distributed satellites, even with the measurement noise unchanged.

\begin{figure*}
    \centering
    \includegraphics[trim=0 0 0 0,clip,width=1.6\columnwidth]{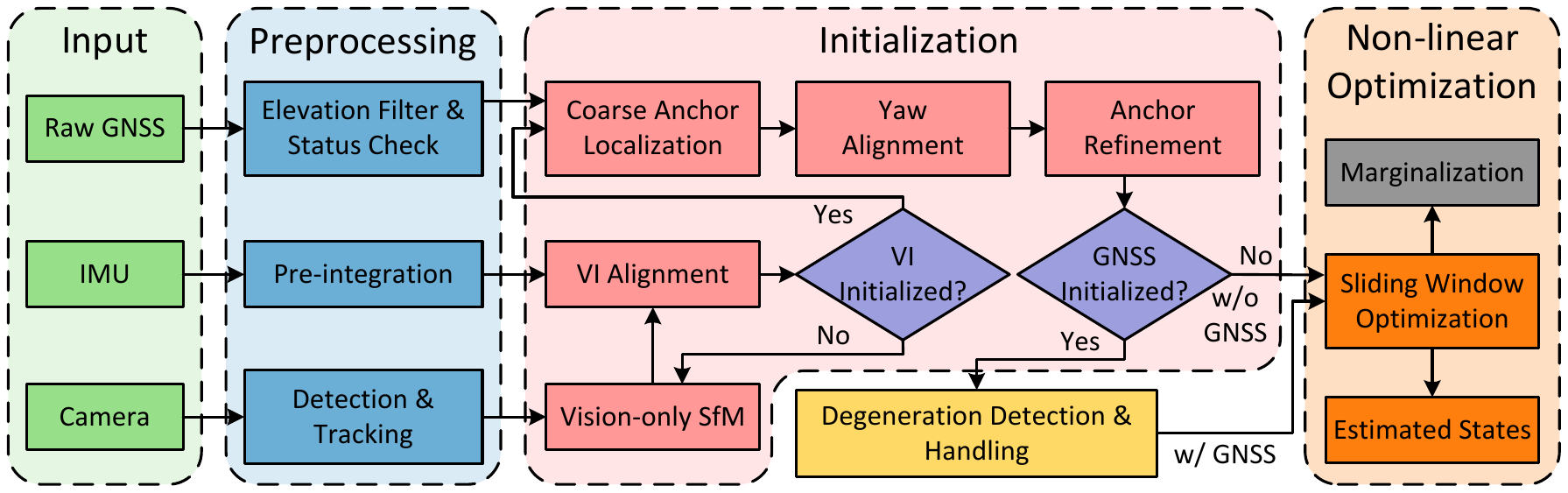}
    \caption{
        \label{fig:system_diagram} 
        The diagram above shows the workflow of our proposed system. At first measurements from all sensors are preprocessed before going into follow-up procedures. In the initialization stage, visual-inertial initialization is accomplished by aligning the inertial information with the result of vision-only SfM. If visual-inertial successfully gets aligned, a coarse-to-fine process is performed in order to initialize the GNSS states. The system monitors and handles GNSS degeneration cases once GNSS states get involved. Finally constraints from all measurements within the sliding window are optimized by the non-linear optimization. Note that if GNSS states cannot get initialized, our system can still work in visual-inertial mode. The marginalization strategy is also adopted to ensure real-time estimation.}
\end{figure*}

\section{System Overview}
\label{sec:system_overview}
The structure of our proposed system is illustrated in Fig.~\ref{fig:system_diagram}. The estimator takes raw GNSS, IMU and camera measurements as input, and applies necessary preprocessing on each type of measurement afterwards. As in \cite{qin2018vins}, the IMU measurements are pre-integrated and sparse feature points are detected and tracked from the image sequence. For GNSS raw data, we first filter out low-elevation and unhealthy satellites which are prone to errors. In order to reject unstable satellite signal, only satellites which are continuously locked for a certain amount of epochs are allowed to enter the system. Because the ephemeris data is acquired via the slow satellite-receiver wireless link (50 bit/s on GPS L1C), a GNSS measurement is unusable until its corresponding ephemeris is fully transmitted. After the preprocessing phase, all measurements are ready for the estimator. Before performing optimization, an initialization phase is necessary to properly initialize the system states of the non-linear estimator.

The initialization starts with a vision-only \ac{SfM}, from which an up-to-similarity motion and structure are jointly estimated, then the trajectory from IMU is aligned to the \ac{SfM} result in order to recover the scale, velocity, gravity and IMU bias. After VI initialization is finished, a coarse-to-fine GNSS initialization process is conducted. At first a coarse anchor localization result is obtained by the SPP algorithm, then the local and global frames are associated in the yaw alignment stage using the local velocity from VI initialization and GNSS Doppler measurement. Finally the initialization phase ends with the anchor refinement, which utilizes accurate local trajectory and imposes clock constraints to further refine the anchor's global position. 

After the initialization phase, the GNSS degeneration cases are checked and carefully handled to ensure robust performance. Then constraints from all measurements are formulated to jointly estimate system states within the sliding window under the non-linear optimization framework. Note that our system is naturally degraded to a VIO if GNSS is not available or cannot be properly initialized. To ensure the real-time performance and handle visual-inertial degenerate motions, the two-way marginalization strategy\cite{SheMicKum1505}, which selects the frame to remove based on a parallax test, is also applied after each optimization.

\section{Probabilistic Formulation}
\label{sec:formulation}
In this section, we first formulate and derive our state estimation problem under the probabilistic framework. As shown later, the whole problem is organized as a factor graph and measurements from sensors form a series of factors which in turn constrain the system states. Each type of factor in the probabilistic graph will be discussed in detail through this section. Note that the formulation of visual and inertial factors are inherited from \cite{qin2018vins} \cite{forster2017manifold} \cite{7139939} thus not the contribution of this work. The relevant content is listed only for the completeness of this literature.

\subsection{MAP Estimation}
We define the optimum system state as the one that maximizes a posterior (MAP) given all the measurements. Assuming that all measurements are independent to each other and the noise with each measurement is zero-mean Gaussian distributed, the MAP problem can be further transformed to the one that minimize the sum of a series of costs, with each cost corresponding to one specific measurement. 
\begin{equation}
	\label{eq:formulation}
	\begin{split}
		\mathcal{X}^{\star}    &= \argmaxA_{\mathcal{X}} p(\mathcal{X} | \mathbf{z}) \\
		&= \argmaxA_{\mathcal{X}} p(\mathcal{X}) p({\mathbf{z} | \mathcal{X}})  \\
		&= \argmaxA_{\mathcal{X}} p(\mathcal{X}) \prod^{n}_{i=1} p({\mathbf{z}_{i} | \mathcal{X}})  \\
		&= \argminA_{\mathcal{X}} \left \{ \left 
		\| \mathbf{r}_p - \mathbf{H}_p \mathcal{X} \right \|^2 
		+ \sum_{i=1}^{n}  \left \| \mathbf{r}(\mathbf{z}_{i},\, \mathcal{X}) \right \|_{\mathbf{P}_{i}}^2\right\} ~,
	\end{split}
\end{equation}
where $\mathbf{z}$ stands for the aggregation of n independent sensor measurements and $\{\mathbf{r}_p, \mathbf{H}_p\}$ encapsulates the prior information of the system state. $\mathbf{r}(\cdot)$ denotes the residual function of each measurement and $\|\cdot\|_{\mathbf{P}}$ is the Mahalanobis norm.

\begin{figure}
    \centering
    \includegraphics[width=0.9\columnwidth]{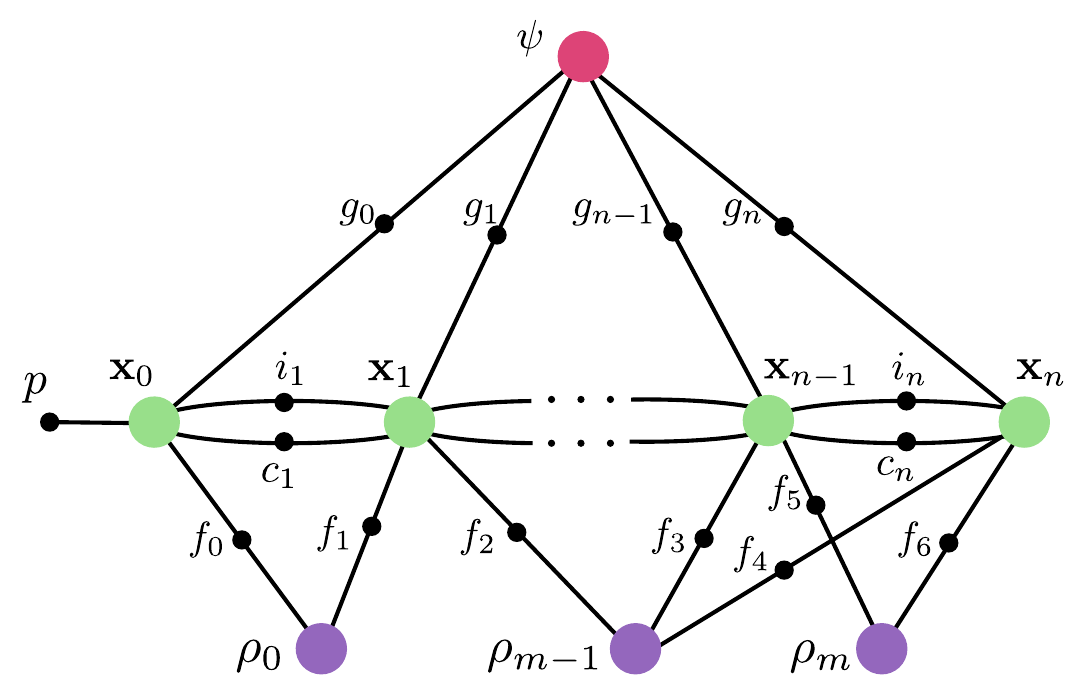}
    \caption{
        \label{fig:factor_graph} 
		Factor graph representation of the optimization problem in our system, where system states are denoted by large colored circles and factors are represented by small black circles. The factors from various measurements consist of inertial factor $i$, visual factor $f$, code pseudorange and Doppler factor $g$ and clock factor $c$. A prior factor $p$ is used to constrain the first pose of the local world frame.}
	% \vspace{-0.3cm}
\end{figure}

Note that such formulation naturally fits with the factor graph representation\cite{910572}, thus we decompose our optimization problem as individual factors that relate states and measurements. Fig.~\ref{fig:factor_graph} shows the factor graph of our system. Besides factors derived from measurements, a prior factor is used to constrain the four unobservable directions of the initial pose of the local world frame, and later it will become a densely connected prior as we marginalize old frames. In the following we will discuss each factor in details.

\subsection{Inertial Factor}
The measurements involved in the inertial factor consist of the biased, noisy linear acceleration and angular velocity of the platform. As the accelerometer operates near the Earth's surface, the linear acceleration measurement also contains the gravity component. The Coriolis and centrifugal forces due to Earth's rotation are ignored in the IMU's formulation considering the noisy measurement of the low-cost IMU. Thus the inertial measurement can be modelled as

\begin{subequations}
	\begin{align}
		\tilde{\mathbf{a}}_t &=  {\mathbf{a}}_t + \mathbf{b}_{a_t} + {\mathbf{R}^{b_t}_w} \mathbf{g}^w + \mathbf{n}_a \\
		\tilde{\boldsymbol{\omega}}_t &=  {\boldsymbol{\omega}}_t + \mathbf{b}_{w_t} + \mathbf{n}_w ~,
	\end{align}
\end{subequations}
where $\{ \tilde{\mathbf{a}}_t,~ \tilde{\boldsymbol{\omega}}_t \}$ is the output of the IMU at time $t$, and $\{ \mathbf{a}_t,~ \boldsymbol{\omega}_t \}$ stands for the linear acceleration and angular velocity of the platform in IMU sensor frame. The additive noise $\mathbf{n}_a$ and $\mathbf{n}_w$ are assumed to be zero-mean Gaussian distributed, e.g., $ \mathbf{n}_a \sim \mathcal{N}(\mathbf{0}, \boldsymbol{\Sigma}_a)$, $ \mathbf{n}_w \sim \mathcal{N}(\mathbf{0},\boldsymbol{\Sigma}_w)$. The slowly varying biases associated with the accelerometer and gyroscope are modelled as a random walk as follows:
\begin{equation}
	\dot{\mathbf{b}}_{a_t} = \mathbf{n}_{b_a},\quad \dot{\mathbf{b}}_{w_t} = \mathbf{n}_{b_w} ~,
\end{equation}
with $ \mathbf{n}_{b_a} \sim \mathcal{N}(\mathbf{0},\boldsymbol{\Sigma}_{b_a})$, $ \mathbf{n}_{b_w} \sim \mathcal{N}(\mathbf{0},\boldsymbol{\Sigma}_{b_w})$.

In practice, the frequency of IMU is often an order of magnitude higher than that of camera, thus it is computationally intractable to estimate each state of the IMU measurements. To this end, IMU pre-integration approach\cite{forster2017manifold} is adopted to aggregate multiple measurements into a single one. For inertial measurements within the time interval $[t_k, t_{k+1}]$, the derived measurements are computed as 
\begin{subequations}
	\label{eq:imu_pre_integration}
	\begin{align}
		\boldsymbol{\alpha}^{b_{t_k}}_{b_{t_{k+1}}} &= \iint_{t \in [t_k,t_{k+1}]} \mathbf{R}^{b_{t_k}}_{b_t} (\tilde{\mathbf{a}}_{t} - \mathbf{b}_{a_t}) dt^2 \\
		\boldsymbol{\beta}^{b_{t_k}}_{b_{t_{k+1}}}  &=  \int_{t \in [t_k,t_{k+1}]} \mathbf{R}^{b_{t_k}}_{b_t} (\tilde{\mathbf{a}}_{t} - \mathbf{b}_{a_t}) dt\\ 
		\boldsymbol{\gamma}^{b_{t_k}}_{b_{t_{k+1}}} &=\int_{t \in [t_k,t_{k+1}]} \frac{1}{2} \boldsymbol{\Omega}(\tilde{\boldsymbol{\omega}}_t - \mathbf{b}_{w_t}) \boldsymbol{\gamma}^{b_{t_k}}_{b_t} dt \, ,
	\end{align}
\end{subequations}
with 
\begin{equation}
	\begin{split}
	\boldsymbol{\Omega}(\boldsymbol{\omega}) = 
	\begin{bmatrix}
	-\lfloor {\boldsymbol{\omega}} \rfloor_{\times} & {\boldsymbol{\omega}}\\
	-{\boldsymbol{\omega}}^T & 0
	\end{bmatrix} ,
	\lfloor \boldsymbol{\omega} \rfloor_{\times} =
	\begin{bmatrix}
	0& -\omega_z & \omega_y\\
	\omega_z& 0 & -\omega_x\\
	-\omega_y& \omega_x & 0
	\end{bmatrix} .
	\end{split}
\end{equation}

Here $ b_k $ stands for the body frame in time $ t_k $. $\{{\boldsymbol{\alpha}}, {\boldsymbol{\beta}}, {\boldsymbol{\gamma}}\}$ encapsulates the relative position, velocity and rotation information between frame $b_k$ and $b_{k+1}$, and can be constructed without the initial position, velocity and rotation profiles given IMU biases. Finally the residual relates the system states and pre-integrated IMU measurements can be formulated as:
\begin{equation}
	\begin{split}
	& \mathbf{r}_{\mathcal{B}}(\tilde{\mathbf{z}}^{b_{t_k}}_{b_{t_{k+1}}},\, \mathcal{X})=
	\begin{bmatrix}
	\delta\boldsymbol{\alpha}^{b_{t_k}}_{b_{t_{k+1}}}\\
	\delta\boldsymbol{\beta}^{b_{t_k}}_{b_{t_{k+1}}}\\
	\delta\boldsymbol{\theta}^{b_{t_k}}_{b_{t_{k+1}}}\\
	\delta{\mathbf{b}_a}\\
	\delta{\mathbf{b}_g}\\
	\end{bmatrix}\\
	&=\begin{bmatrix}
	\mathbf{R}^{b_{t_k}}_{w}(\mathbf{p}^{w}_{b_{t_{k+1}}} - \mathbf{p}^{w}_{b_{t_k}} + \frac{1}{2}\mathbf{g}^{w} \Delta t_k^2 - \mathbf{v}^{w}_{b_{t_k}} \Delta t_k) -\boldsymbol{\hat{\alpha}}^{b_{t_k}}_{b_{t_{k+1}}} \\
	\mathbf{R}^{b_{t_k}}_{w}(  \mathbf{v}^{w}_{b_{t_{k+1}}}   + \mathbf{g}^{w} \Delta t_k- \mathbf{v}^{w}_{b_{t_k}})-  \boldsymbol{\hat{\beta}}^{b_{t_k}}_{b_{t_{k+1}}} \\
	2\begin{bmatrix}\mathbf{q}^{w^{-1}}_{b_{t_k}} \otimes\mathbf{q}^{w}_{b_{t_{k+1}}}
	\otimes \inv{(\hat{\boldsymbol{\gamma}}^{b_{t_k}}_{b_{t_{k+1}}})}
	\end{bmatrix}_{xyz}\\
	{\mathbf{b}_a}_{b_{t_{k+1}}} - {\mathbf{b}_a}_{b_{t_k}}\\
	{\mathbf{b}_w}_{b_{t_{k+1}}} - {\mathbf{b}_w}_{b_{t_k}}
	\end{bmatrix},
	\end{split}
\end{equation}
where $\delta\boldsymbol{\theta}^{b_{t_k}}_{b_{t_{k+1}}}$ represents the relative rotation error in 3D Euclidean space, and the operator $\begin{bmatrix}\cdot \end{bmatrix}_{xyz}$ returns the imaginary part of a quaternion.

\subsection{Visual Factor}
The visual measurement used in our system is a bunch of sparse feature points extracted from image frames. The strong corners\cite{323794} within the image are detected as feature points and are further tracked by the iterative Lucas-Kanade method\cite{LucKan8108}. After distortion correction \cite{6696592} being applied to feature points, the projection process can be modelled as:
\begin{equation}
	\label{eq:projection}
	\begin{split}
		\widetilde{\mathcal{P}} &= \pi_c(\mathbf{R}_{b}^{c} (\mathbf{R}_{w}^{b}\,\mathbf{x}^w + \mathbf{p}_{w}^{b}) + \mathbf{p}_{b}^{c}) + \mathbf{n}_c ~,
	\end{split}
\end{equation}
where $\widetilde{\mathcal{P}} = [u, ~v]^T$ is the feature coordinate in image plane, and $\mathbf{x}^w$ is its corresponding 3D landmark position in local world frame. $\pi_c(\cdot)$ represents the camera projection function and $\mathbf{n}_c$ is the measurement noise. Thus for a feature $l$ with inverse depth $\rho_{l}$ in frame $i$, if it is observed again in frame $j$, the residual that relates two frames can be expressed as 

\begin{subequations}
	\label{eq:visual_residual}
	\begin{align}
		&\mathbf{r}_{\mathcal{C}}(\tilde{\mathbf{z}}_l,\, \mathcal{X}) = \widetilde{\mathcal{P}}^{c_{t_j}}_{l} - \pi_c(\hat{\mathbf{x}}^{c_{t_j}}_{l}) \\
	\begin{split}
		&\hat{\mathbf{x}}^{c_{t_j}}_{l} = 
		\mathbf{R}^c_b 
		(\mathbf{R}^{b_{t_j}}_w
		(\mathbf{R}^w_{b_{t_i}}
		(\mathbf{R}^b_c 
		\frac{1}{\rho_l} 
		\inv{\pi_c} (\widetilde{\mathcal{P}}^{c_{t_i}}_{l}) + \mathbf{p}^b_c) + \\
		& \quad \quad \quad \quad \mathbf{p}^w_{b_{t_i}}) + \mathbf{p}_w^{b_{t_j}}) + \mathbf{p}_b^c ~, 
	\end{split}
	\end{align}
	\end{subequations}
where $\{ \mathbf{R}_c^b, \mathbf{t}_c^b\}$ is the transformation between IMU and camera.

\subsection{Code Pseudorange Factor}
Consider a GNSS receiver $r$ which locks a navigation satellite $s$, it measures the code shift to obtain the code pseudorange information as illustrated in Eq.~\eqref{eq:code_pseudorange}. The satellite clock error and atmospheric delay are compensated using the models described in Section \ref{ssec:spp_algorithm}. In our system, the code pseudorange noise $\epsilon_{r}^{s}$ is assumed to be zero-mean Gaussian distributed such as $\epsilon_{r}^{s} \sim N(0, \sigma_{r, pr}^{s}) $, where the variance $\sigma_{r, pr}^{s}$ is modelled as 
\begin{equation}
	\label{eq:pr_noise}
	\begin{split}
		\sigma_{r, pr}^{s} &= \frac{n_{s} \times n_{pr}}{\sin^{2} \theta_{el}} ~.
	\end{split}
\end{equation}
Here $n_{s}$ is the broadcast satellite space accuracy index, and $n_{pr}$ is the code pseudorange measurement noise index reported by the receiver. $\theta_{el}$ represents the satellite elevation angle at the view of the receiver, and there are two reasons for this denominator term. Firstly it can suppress the noise caused by GNSS multiple path effect that usually occurs on low elevation satellites. Furthermore, the ionospheric delay obtained by Klobuchar model, which is widely adopted by navigation system, still contains an error up to $50 \%$\cite{klobuchar1987ionospheric}. As the low elevation satellites will suffer from a significant ionospheric delay, the denominator term can also reduce the error coming with the ionospheric compensation.

A coordinate in the ECEF frame can be transformed to the local world frame via an anchor point, at which the ENU frame is built. Given the ECEF coordinate of the anchor point, the rotation from ENU frame to ECEF frame is 
\begin{equation}
	\label{eq:R_ecef_enu}
	\begin{split}
		\mathbf{R}_{n}^{e} &= \begin{bmatrix}
			-\sin \lambda \!&\! -\sin \phi \, \cos \lambda \!&\!  \cos \phi \, \cos \lambda \\
			\quad \cos \lambda \!&\! -\sin \phi \, \sin \lambda \!&\!  \cos \phi \, \sin \lambda \\
			\quad 0 \!&\! \cos \phi \!&\!  \sin \phi
		   \end{bmatrix}\! ~,
	\end{split}
\end{equation}
where $\phi$ and $\lambda$ is the latitude and longitude of the reference point in geographic coordinate system. The 1-DOF rotation between ENU and local world frame $\mathbf{R}_{w}^{n}$ is given by the yaw offset $\psi$. Then the relationship between ECEF and local world coordinates of the receiver's antenna can be expressed as 
\begin{equation}
	\label{eq:local2ecef}
	\begin{split}
		\mathbf{p}^{e}_{r} &= \mathbf{R}_{n}^{e} \mathbf{R}_{w}^{n} (\mathbf{p}^{w}_{r} - \mathbf{p}^{w}_{anc}) + \mathbf{p}^{e}_{anc} ~.
	\end{split}
\end{equation}

In our implementation we set the anchor point to the origin of the local world frame, that is, the origin of the local world frame coincides with the origin of the ENU frame, as illustrated in Fig.~\ref{fig:coordinate_systems}. Thus $\mathbf{p}^{w}_{anc}$, the anchor's coordinate in the local world frame, becomes a zero vector. The position of the receiver's antenna in the local world frame can be associated with the system states by 
\begin{equation}
	\label{eq:body2rcv}
	\begin{split}
		\mathbf{p}^{w}_{r} &= \mathbf{p}^{w}_{b} + \mathbf{R}^{w}_{b} \mathbf{p}^{b}_{r}  ~,
	\end{split}
\end{equation}
where $\mathbf{p}^{b}_{r}$ is the offset of the antenna expressed in body frame. 

So far we are able to compute the ECEF coordinate of the receiver's antenna at any time given the corresponding system states. Because the GNSS measurements are time tagged by the receiver, we define the ECI frame to be coincident with the ECEF frame at the signal reception time. In this way, we have $\mathbf{p}^{E}_{r} = \mathbf{p}^{e}_{r}$ when the signal arrives at the receiver. On the other hand, the satellite's position in ECEF frame at the signal transmission time, which we denote as $\mathbf{p}^{e'}_{s}$, can be obtained by the broadcast ephemeris and code pseudorange measurement. As a result of Earth's rotation, the ECEF frame $(\cdot)^{e'}$ when the signal leaves the satellite is different from the one $(\cdot)^{e}$ when the signal arrives. To this end, the satellite's position need to be transformed to the ECI frame (also the ECEF frame at reception time) by
\begin{equation}
	\label{eq:sat_ecef2eci}
	\begin{split}
		\mathbf{p}^{E}_{s} &= \mathbf{R}_{z}\big(-{\omega}_E \, t_f\big) ~ \mathbf{p}^{e'}_{s}  ~,
	\end{split}
\end{equation}
where $\mathbf{R}_{z}(\theta)$ represents a rotation about the $z$ axis of the ECI frame with magnitude $\theta$, and $t_f$ is the \ac{ToF} of the GNSS signal.

In the end, the residual of a single code pseudorange measured in $t_k$, which connects system states ${ \{ \mathbf{p}^w_{b_{t_k}}, \mathbf{q}^w_{b_{t_k}}, \delta \mathbf{t}_k, \psi \}}$ and satellite $s_j$, can be formulated as
\begin{equation}
	\label{eq:pr_cost}
	\begin{split}
		r_{\mathcal{P}}(\tilde{\mathbf{z}}_{r_k}^{s_j}, \, \mathcal{X})
		% r_{r, pr}^{s} 
		= & \| \mathbf{R}_{z}({\omega}_E \, t_f) ~ \mathbf{p}^{e'}_{s} - \mathbf{p}^{E}_{r_k} \| + c (\bm{\zeta}_{s_j}^T \delta \mathbf{t}_k - \Delta t^{s_j}) + \\
		& T_{r_k}^{s_j} + I_{r_k}^{s_j} - \tilde{P}_{r_k}^{s_j} ~,
	\end{split}
\end{equation}

where $r_k$ stands for the GNSS receiver at time $t_k$.

% R_ecef_enu * d_yaw * local_pos
\subsection{Doppler Factor}
The Doppler frequency shift, as shown in Eq.~\eqref{eq:Doppler}, is a result of the relative velocity along the line of the signal propagation path between the receiver and satellite. Similar to code pseudorange noise, the Doppler measurement noise $\eta_{r,dp}^{s}$ is assumed to be Gaussian distributed and the corresponding variance is modelled as 
\begin{equation}
	\label{eq:dp_noise}
	\begin{split}
		\sigma_{r,dp}^{s} &= \frac{n_{s} \times n_{dp}}{\sin^{2} \theta_{el}} ~,
	\end{split}
\end{equation}
where $n_{dp}$ is the measurement noise index reported by the receiver. The receiver's velocity in ECEF frame can be obtained from the local world velocity via 

\begin{equation}
	\label{eq:vel_local2ecef}
	\begin{split}
		\mathbf{v}^{e}_{r} &= \mathbf{R}_{n}^{e} \mathbf{R}_{w}^{n} \mathbf{v}^{w}_{b} ~.
	\end{split}
\end{equation}

By defining the ECI frame as the ECEF frame at reception time, we have $\mathbf{v}^{E}_{r} = \mathbf{v}^{e}_{r}$. Then the satellite's velocity in the signal-transmission ECEF frame, $\mathbf{v}^{e'}_{s}$, can be transformed to the ECI frame by 
\begin{equation}
	\label{eq:sat_vel_ecef2eci}
	\begin{split}
		\mathbf{v}^{E}_{s} &= \mathbf{R}_{z}\big(-{\omega}_E \, t_f\big) ~ \mathbf{v}^{e'}_{s}  ~,
	\end{split}
\end{equation}

 Finally the residual with related to Doppler measurement in $t_k$, which connects system states ${ \{ \mathbf{p}^w_{b_{t_k}}, \mathbf{v}^w_{b_{t_k}}, \dot{\delta t}_k, \psi \}}$ and satellite $s_j$, can be formulated as 
\begin{equation}
	\label{eq:dp_cost}
	\begin{split}
		r_{\mathcal{D}}(\tilde{\mathbf{z}}_{r_k}^{s_j}, \, \mathcal{X})
		% r_{r, dp}^{s} 
		= & \frac{1}{\lambda} {\bm{\kappa}_{r_k}^{s_j}}^{T} (\mathbf{v}^{E}_{s_j} - 
		\mathbf{v}^{E}_{r_k}) + \\
		& \frac{c}{\lambda} (\dot{\delta t}_k - \dot{\Delta t^{s_j}}) + \Delta \tilde{f}^{s_j}_{r_k} ~.
	\end{split}
\end{equation}

\subsection{Receiver clock factors}
The receiver clock biases in $t_k$ and $t_{k+1}$ relate the clock drift rate by 
\begin{equation}
	\label{eq:dt_ddt_factor}
	\begin{split}
		\delta \mathbf{t}_k &= \delta \mathbf{t}_{k-1} + \mathbf{1}_{4 \times 1} \int_{t_{k-1}}^{t_k} 
		\dot{\delta t} \,dt ~,
	\end{split}
\end{equation}
where $ \mathbf{1}_{n \times m} $ stands for $n$ by $m$ all-ones matrix, and the residual in discrete case is 
\begin{equation}
	\label{eq:dt_ddt_cost}
	\begin{split}
		\mathbf{r}_{\mathcal{T}}(\tilde{\mathbf{z}}_{k-1}^{k}, \, \mathcal{X})
		% \mathbf{r}_{t, k} 
		&= \delta \mathbf{t}_k - \delta \mathbf{t}_{k-1} - \mathbf{1}_{4 \times 1} \dot{\delta t}_{k-1} \tau_{k-1}^{k} ~,
	\end{split}
\end{equation}
where $ \tau_{k-1}^{k} $ is the time difference between measurement $ k\!-\!1 $ and $ k $. The covariance matrix associate with this residual is defined as a $4$ by $4$ diagonal matrix $\mathbf{D}_{t, k}$ with its elements describe the discretization error.

The GNSS receiver clock drift rate, on the other hand, is determined by the frequency stability of the receiver clock. \ac{TCXO} is often chosen as the clock source on low-cost GNSS receivers. Due to the noise characteristic of \ac{TCXO}, the receiver clock drift rate is modelled as a random walk process, thus the residual becomes 
\begin{equation}
	\label{eq:ddt_smooth_cost}
	\begin{split}
		r_{\mathcal{W}}(\tilde{\mathbf{z}}_{k-1}^{k}, \, \mathcal{X})
		% r_{dt, k} 
		&= \dot{\delta t}_k - \dot{\delta t}_{k-1} ~.
	\end{split}
\end{equation}
The corresponding variance $ \sigma_{dt, k} $ is determined by the stability of the clock frequency drift.

\section{GNSS Initialization and Degeneration}
\label{sec:initialization_and_degeneration}
The state estimation process described in the last section is non-linear with respect to the system states thus its performance heavily relies on the initial values. With online initialization, the initial states can be well recovered from an unknown situation without any assumption or manual intervention. During the system operation, the estimator may also encounter imperfect situations where some of sensors experience failure or degeneration. As there is already extensive literature on the topics of initialization and degeneration with respect to the visual-inertial system, in this section we limit the scope to the GNSS part. In the following we first introduce the proposed coarse-to-fine GNSS initialization approach, then we discuss several scenarios that degrade the performance of our system.

\begin{figure}
    \centering
    \includegraphics[width=1.0\columnwidth]{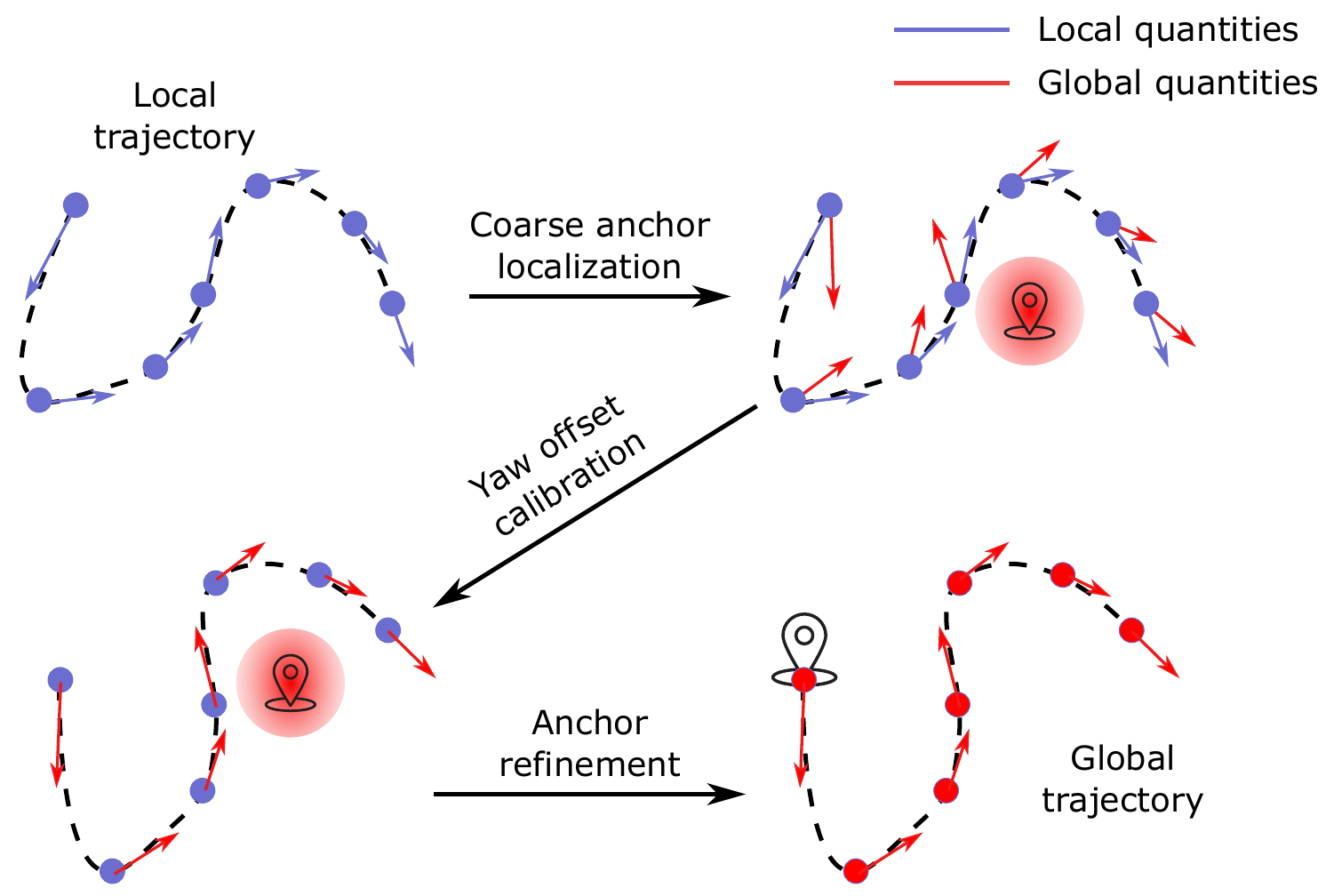}
    \caption{
        \label{fig:initialization} 
        An illustration of the proposed coarse-to-fine initialization process. The module takes the local position and velocity result from VIO and outputs the corresponding trajectory in global ECEF frame.}
	% \vspace{-0.3cm}
\end{figure}

\subsection{Initialization}
As mentioned before, an anchor point with known global and local coordinate is necessary to fuse the global GNSS measurement with the local visual and inertial information. As the anchor point is already set to the origin of the local world frame, the ECEF coordinate of the local world origin need to be calibrated beforehand. In addition, the yaw offset $\psi$ between ENU and local world frame, which brings nonlinearity into the system, also needs a reasonable initial value in order to converge at the non-linear optimization stage. In this paper, we propose a multi-stage GNSS-VI initialization procedure to online calibrate the anchor point and the yaw offset. Before the GNSS-VI initialization, we assume that the VIO has been successfully initialized, i.e. the gravity vector, initial velocity, initial IMU bias and scale have obtained initial values\cite{QinShen17}. After that, a smooth trajectory in the local world frame is formed and is ready to be used in the GNSS-VI initialization phase. The GNSS-VI initialization procedure requires at least 4 satellites being tracked (if all satellites belongs to a single system, $(N+3)$ if $N$ satellite systems are involved). In addition, a minimum distance of 4 meters is also required to obtain reliable initial quantities. As illustrated in Fig.~\ref{fig:initialization}, the online GNSS-VI initialization is conducted in a coarse-to-fine manner and consists of three steps:

\subsubsection{Coarse Anchor Point Localization}
At first a coarse ECEF coordinate is generated by the GNSS SPP algorithm without any prior information. The SPP algorithm takes all code pseudorange measurements from the most recent epoch as input. 

\subsubsection{Yaw Offset Calibration}
In the second step, we calibrate the yaw offset between the ENU frame and the local world frame using the less noisy Doppler measurement. The initial yaw offset and receiver clock drift rate are obtained through the following optimization problem
\begin{equation}
	\label{eq:yaw_initialize}
	\begin{split}
		\underset{\dot{\delta t}, \psi}{\text{minimize}} \:  
			\sum_{k=1}^{n} \sum_{j=1}^{p_k}
			\left \| r_{\mathcal{D}}(\tilde{\mathbf{z}}_{r_k}^{s_j}, \, \mathcal{X}) \right \|_{\sigma_{r_k, dp}^{s_j}} ^{2} ~,
	\end{split}
\end{equation}
where n is the sliding window size and $ p_k $ is the number of satellites observed in $k$-th epoch inside the window. Here we fix the velocity $ \mathbf{v}_b^w $ to the result of VIO and assume that $ \dot{\delta t}_k $ is constant within the window. The coarse anchor coordinate obtained from the first step is used to calculated the direction vector $\bm{\kappa}_{r}^{s}$ and rotation $\mathbf{R}_{n}^{e}$. $\bm{\kappa}_{r}^{s}$ and $\mathbf{R}_{n}^{e}$ are not sensitive to the receiver's location thus a coarse anchor point coordinate is sufficient. The parameters to be estimated only include the yaw offset $\psi$ and the average clock bias drift rate $\dot{\delta t}$ over the entire window measurements. After that, the transformation between the ENU frame and local world frame is fully calibrated.
\subsubsection{Anchor Point Refinement}
Finally we are ready to refine the previous coarse anchor point and align the local world trajectory with that in ECEF frame. Different from the first step, the position result from VIO is used as prior information. The following problem is optimized over the sliding window measurements.
\begin{equation}
	\label{eq:refine_anchor}
	\begin{split}
		\underset{\mathbf{\delta t}, \, \mathbf{p}_{anc}^{e}}{\text{minimize}} \:  
			& \Big( \sum_{k=1}^{n} \sum_{j=1}^{p_k} \left \|
			r_{\mathcal{P}}(\tilde{\mathbf{z}}_{r_k}^{s_j}, \, \mathcal{X}) \right \| _{\sigma_{r_k, pr}^{s_j}}^{2} + \\
			& \sum_{k=1}^{n} 
				\left \| \mathbf{r}_{\mathcal{T}}(\tilde{\mathbf{z}}_{k-1}^{k}, \, \mathcal{X}) \right \|_{\mathbf{D}_{t, k}}^{2} \Big)
	\end{split}
\end{equation}

The anchor point coordinate and the receiver clock biases associate with each GNSS epoch are refined through the optimization of the above problem. After this step, the anchor point, origin of the ENU frame, is set to the origin of the local world frame. Finally the initialization phase of the entire estimator is finished and all necessary quantities have been assigned initial values.

\subsection{Degenerate Cases}
There is no doubt that our fusion system will perform best in an open-area where GNSS signal is stable and satellites are well-distributed. In the following we will discuss several situations which may degrade the performance of our system.
\subsubsection{Low speed movement} %%% CITE paper for an order of magnitude lower
Since the noise level of Doppler shift measurement is an order of magnitude lower than that of code pseudorange, the yaw offset between the local world frame and ENU frame can be well constrained by a short window of Doppler shift measurements. Once the velocity of the GNSS receiver is below the noise level of the Doppler shift, the estimated yaw offset may be corrupted by the measurement noise. In addition, low speed movement also implies that the translational distance within the window is short, thus the yaw estimation may be affected by code pseudorange as well. In an extreme case where the platform experiences a rotation-only movement, GNSS cannot provide any information on the rotational directions and in turn the yaw component will drift as that in VIO. Thus we fix the yaw offset variable if the average velocity inside the window is below the threshold $v_{ths}$. In our system, $v_{ths}$ is set to $0.3 ~ m/s$ which can be easily satisfied even by a pedestrian.

\subsubsection{Less than 4 satellites being tracked}
If the number of satellites being tracked is less than 4, the SPP or loosely-coupled approaches will fail to resolve the receiver's location. However, with the help of the tightly-coupled structure, our system is still able to make use of available satellites and subsequently update the states vector. Later in Section \ref{ssec:real_world_experiments} we will investigate the performance degradations under various satellite configurations.

\subsubsection{No GNSS signal}
In indoor or cluttered environments where GNSS signal is totally unavailable, the states related to global information, namely the yaw offset $ \psi $, receiver clock bias $ \mathbf{\delta t} $ and drift rate $ \dot{\delta t} $ are no longer observable. However, constraints from Eq.~\eqref{eq:dt_ddt_cost} and \eqref{eq:ddt_smooth_cost} are still kept during the optimization. The clock drift rate of low-cost receivers is quite stable as we found in the receiver stand-still analysis, thus the (near-)optimum clock drift rate is maintained by the constraint from  Eq.~\eqref{eq:ddt_smooth_cost}. Similarly, the receiver bias is propagated by the constraint from  Eq.~\eqref{eq:dt_ddt_cost}, which in turn provides a good initial value when the GNSS signal is reacquired. This mechanism improves the stability of our fusion system when the GNSS signal is intermittent and eliminates the need for re-initialization when signal is lost-and-reacquired. 

\begin{figure}
    \centering
    \includegraphics[width=0.95\columnwidth]{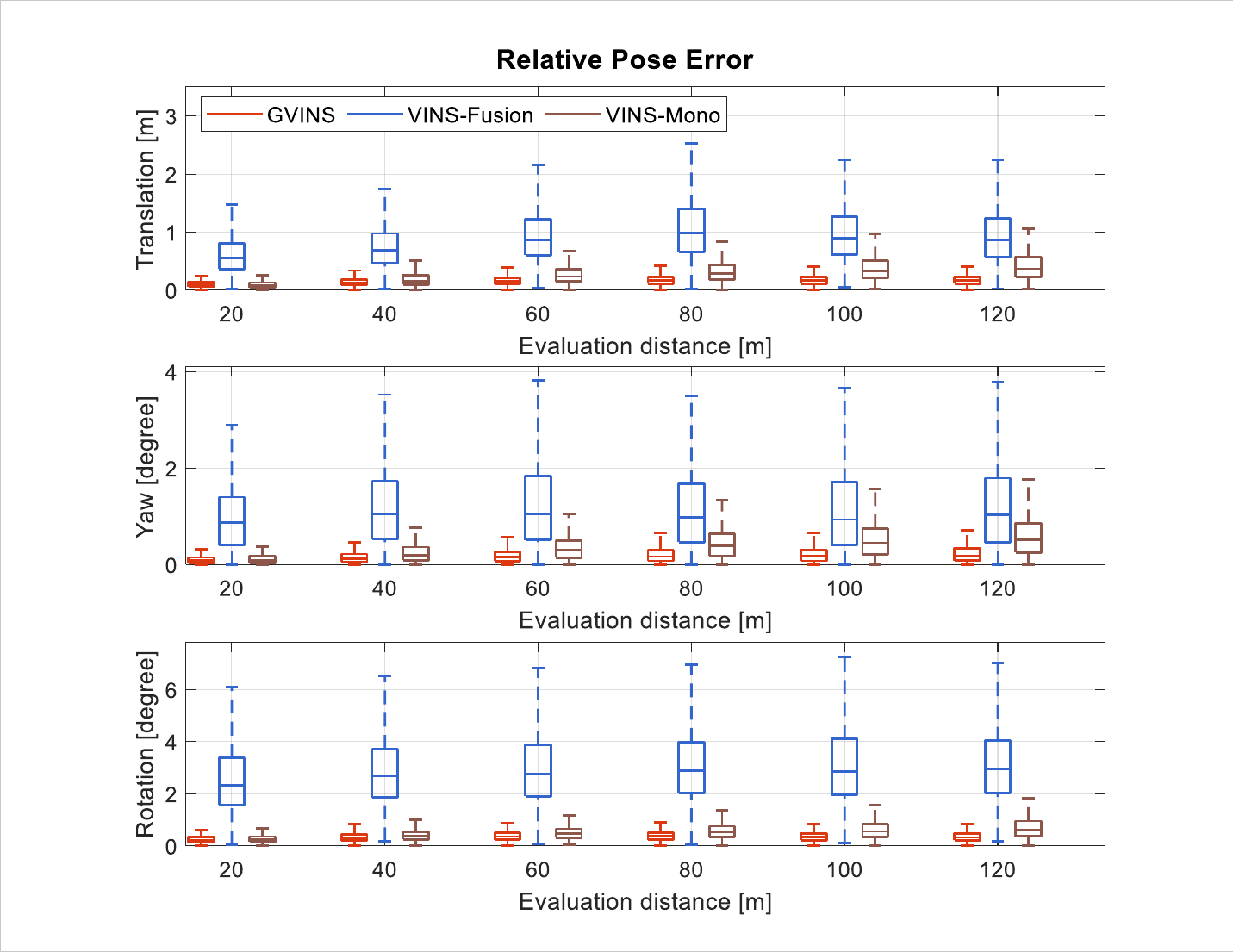}
    \caption{
        \label{fig:simulation_rpe} 
        Relative pose error of GVINS, VINS-Fusion and VINS-Mono with respect to the evaluation distance on the simulation environment. The top two figures correspond to the four unobservable directions (x, y, z and yaw) of VIO and the bottom figure is the overall relative rotation error.}
\end{figure}

\section{Experimental Results}
\label{sec:experiments}
We conduct both simulation and real-world experiments to verify the performance of our proposed system. In this section, we compare our system with the open-source VINS-Mono\cite{qin2018vins}, VINS-Fusion\cite{qin2019general} (Monocular+IMU+GNSS) and RTKLIB\cite{takasu2009development}. Since we are only interested in the real-time estimation result, the loop function of VINS-Mono and VINS-Fusion, which optimizes pose graph based on revisited scene, is disabled. We use RTKLIB\footnote{https://github.com/tomojitakasu/RTKLIB/tree/rtklib\_2.4.3} to compute the GNSS SPP solution and feed the obtained GNSS location to VINS-Fusion for a loose-coupled result. The window size of our system, as well as that of VINS-Mono and VINS-Fusion, are set to 10. Table.~\ref{tab:data_profile} lists the maximum velocity and the overall RTK fixed rate in each experiment. All experiments in this section are performed on a desktop PC with an Intel i7-8700K at 3.7 GHz and 32 GB memory.

\begin{table}[h]
    \centering
	\caption{ Velocity and RTK fixed rate profiles in each experiment \label{tab:data_profile}}
	\setlength\tabcolsep{4pt}
    \begin{tabular}{ccc}
		\toprule
		 & maximum velocity [m/s]  & RTK fixed rate [$\%$] \\
		\midrule
		Simulation & 10.000 & N/A \\
		Sports field & 1.676  & 100 \% \\
		Indoor-outdoor & 2.108 & 81.3 \% \\
		Urban driving & 21.424 & 84.7 \% \\
        \bottomrule
    \end{tabular}
\end{table}

\subsection{Simulation}
\subsubsection{Setup}
The simulation environment is a $ 30m \times 30m \times 30m $ cube with random generated 3D landmarks. The landmarks are projected to a 10-Hz virtual camera with 75 degree horizontal FOV and 55 degree vertical FOV, which in turn generates around 100 visible features per frame. An additional white noise term with a standard deviation of 0.5 pixel is added to all feature points. A virtual 200-Hz IMU is rigidly connected to the camera and moves along a pre-designed 3D path. The standard deviation associate with the white noise of the accelerometer and gyroscope is set to $ 0.05 m/s^2 $ and $ 0.005 rad/s $ respectively, and the standard deviation of the accelerometer and gyroscope bias random walk is set to $ 3.5 \times 10^{-4} m/s^2 $ and $ 3.5 \times 10^{-5} rad/s $ respectively. In the meantime, a 10-Hz virtual GNSS receiver generates code pseudorange and Doppler shift measurements using the past or real-time broadcast ephemeris data. The standard deviation of code pseudorange and Doppler white noise shift is set to $ 1m $ and $ 0.5 $ Hz ($ \sim 0.1m/s $  equivalent) respectively. The simulation experiment lasts for 30 minutes, with a trajectory over 10 kilometers.

\begin{figure}
    \centering
    \includegraphics[width=0.95\columnwidth]{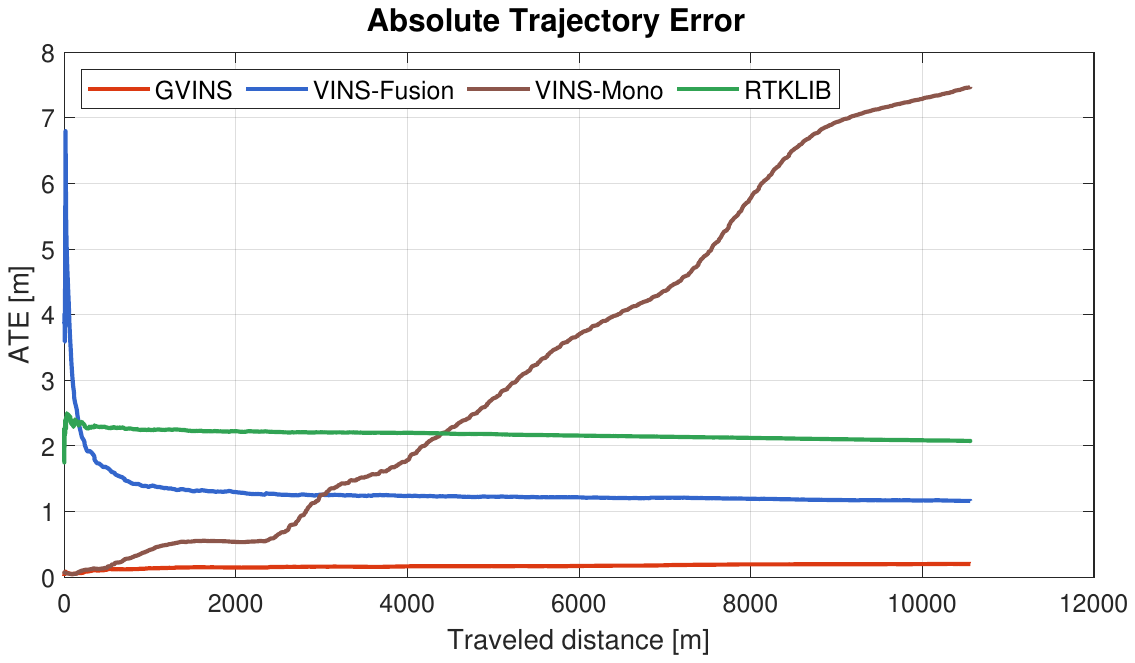}
    \caption{
        \label{fig:simulation_ate} 
        Absolute trajectory error of GVINS, VINS-Fusion, VINS-Mono and RTKLIB with respect to the traveled distance on the simulation environment.}
	% \vspace{-0.3cm}
\end{figure}

\begin{table}[h]
    \centering
	\caption{Initialization quality metrics in simulation and real-world experiments \label{tab:initialization_table}}
	\setlength\tabcolsep{4pt}
    \begin{tabular}{ccc}
		\toprule
		 & Yaw offset error [degree] & Anchor point error [m]  \\
		\midrule
		Simulation & 0.183 & 0.635 \\
		Sports field & 0.35 & 1.491 \\
		Indoor-outdoor & 0.478 & 4.370 \\
		Urban driving & 2.490 & 4.816 \\
        \bottomrule
    \end{tabular}
\end{table}

\subsubsection{result}
In the simulation environment, the GNSS-VI gets initialized immediately after visual-inertial alignment since the system do not need to wait for ephemerides. The initialization quality, which we measure by the error of local-ENU yaw offset and anchor point, is listed in Table.~\ref{tab:initialization_table}. Fig.~\ref{fig:simulation_rpe} shows the relative pose error (RPE)\cite{geiger2012we} with respect to the evaluation distance. As can be seen from the figure, The relative error of VINS-Mono increases with the evaluation distance in both translational and rotational directions. Among those the rotational error mainly comes from yaw component. This indicts that VINS-Mono suffers from accumulated drift in the four unobservable directions, namely x, y, z and yaw. The error of VINS-Fusion exhibits similar tendency when the evaluation distance is short, and remains at a constant level when the distance increases further. This implies that VINS-Fusion is able to bound the accumulated drift by loosely incorporating the GNSS solution. However, the magnitude of its relative error is much larger compared with the result of VINS-Mono and GVINS, thus the smoothness of the estimator is highly affected by the noisy GNSS measurement. Thanks to the tightly-coupled approach we adopted, our proposed system combines advantages of both VINS-Mono and VINS-Fusion. On the one hand, the relative error is comparable to that of VINS-Mono for short range thus the smoothness is preserved. On the other hand, the error no longer accumulates in all directions and the global consistency is also guaranteed.

Fig.~\ref{fig:simulation_ate} depicts the absolute trajectory error (ATE) along with the traveled distance. The error plot of VINS-Mono keeps increasing as a result of accumulated drift, while it remains constant for all other three approaches. The ATE of RTKLIB SPP algorithm shows the noise level of the GNSS code pseudorange measurement, and VINS-Fusion is able to reduce the magnitude of ATE by combine the result of VIO in a loosely-coupled manner. By tightly fusing GNSS raw measurements and visual inertial data in a unified framework, our algorithm effectively suppresses the noise of GNSS signal and keeps the ATE at a low level. The final \ac{RMSE} of each approach is shown in Table.~\ref{tab:rmse_table}.

\begin{table}[h]
    \centering
	\caption{RMSE[m] statistics compared with different approaches in simulation environment \label{tab:rmse_table}}
	\setlength\tabcolsep{4pt}
    \begin{tabular}{ccccc}
		\toprule
		 & GVINS & VINS-Fusion & VINS-Mono & RTKLIB \\
		\midrule
		Simulation & 0.202 & 1.162 & 7.471 & 2.076 \\
		Sports field & 0.806 & 2.149 & 8.537 & 2.835 \\
		Indoor-outdoor & 3.700 & 6.905 & 36.651 & 6.036 \\
		Urban driving & 4.508 & N/A & N/A & 11.106 \\
        \bottomrule
    \end{tabular}
\end{table}

\subsection{Real-world Experiments}
\label{ssec:real_world_experiments}

\begin{figure}
    \centering
    \includegraphics[width=0.95\columnwidth]{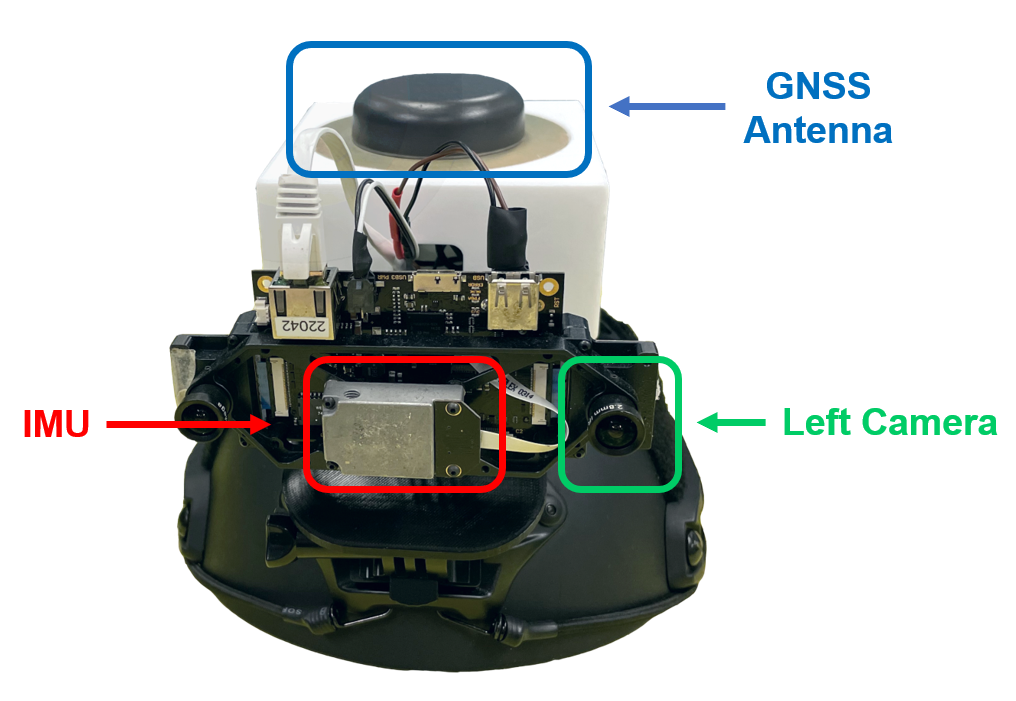}
    \caption{
        \label{fig:sensor_suit} 
        The equipment used in our real-world experiments is a helmet with a VI-Sensor and a u-blox ZED-F9P attached. The camera and IMU measurements are well synchronized by VI-Sensor itself. The PPS signal from the GNSS receiver is used to trigger the VI-Sensor to align the global time with the local time.}
    \vspace{-2.0cm}
\end{figure}

As illustrated in Fig.~\ref{fig:sensor_suit}, the device used in our real-world experiments is a helmet with a VI-Sensor\cite{6906892} and an u-blox ZED-F9P GNSS receiver \footnote{https://www.u-blox.com/en/product/zed-f9p-module} attached. The detailed specifications of each sensor are shown in Table.~\ref{tab:sensor_specs}. The VI-Sensor provides two cameras as a stereo pair and we only use the left one for all experiments. The u-blox ZED-F9P is a low-cost multi-band receiver with multiple constellations support. In addition, ZED-F9P owns an internal RTK engine which is capable to provide receiver's location at an accuracy of 1cm in open area. The real-time RTCM stream from a nearby base station is fed to to the ZED-F9P receiver for the ground truth RTK solution. In terms of time synchronization, the camera and IMU are synchronized by VI-Sensor, and the local time is aligned with the global GNSS time via \ac{PPS} signal of ZED-F9P and hardware trigger of VI-Sensor.

\begin{table}[h]
    \centering
	\caption{Sensor specifications for the device used in real-world environment \label{tab:sensor_specs}}
	\setlength\tabcolsep{4pt}
    \begin{tabular}{ccc}
		\toprule
		\textbf{Sensor Type/Item} & \textbf{Value} & \textbf{Unit} \\
		\midrule
		\textbf{Camera} & & \\
		Sensor & Aptina MT9V034 &  \\
		Shutter & Global shutter & \\
		Resolution & $752 \times 480$ & $ \textrm{pixel} $ \\
		Horizontal field of view & $ 98 $ & $ \textrm{degree} $ \\
		Vertical field of view & $ 73 $ & $ \textrm{degree} $ \\
		Frequency & $20$ & $\textrm{Hz}$ \\
		\midrule
		\textbf{IMU} &  &  \\
		Sensor & ADIS16448 &  \\
		Frequency & $ 200 $ & $\textrm{Hz}$ \\
		Gyroscope noise density & $7.0 \times 10^{-3}$ & $^{\circ}/s ~ \textrm{Hz}^{-0.5}$ \\
		Accelerometer noise density & $6.6 \times 10^{-4}$ & $ ms^{-2} ~ \textrm{Hz}^{-0.5} $\\
		\midrule
		\textbf{GNSS} &  &  \\
		Receiver & u-blox ZED-F9P &  \\
		Antenna & Tallysman TW3882 &  \\
		Raw measurement frequency & $ 10 $ & $\textrm{Hz}$ \\
		RTK solution frequency & $ 10 $ & $\textrm{Hz}$ \\
        \bottomrule
    \end{tabular}
\end{table}

\begin{figure}
	\centering
    \includegraphics[width=0.95\columnwidth]{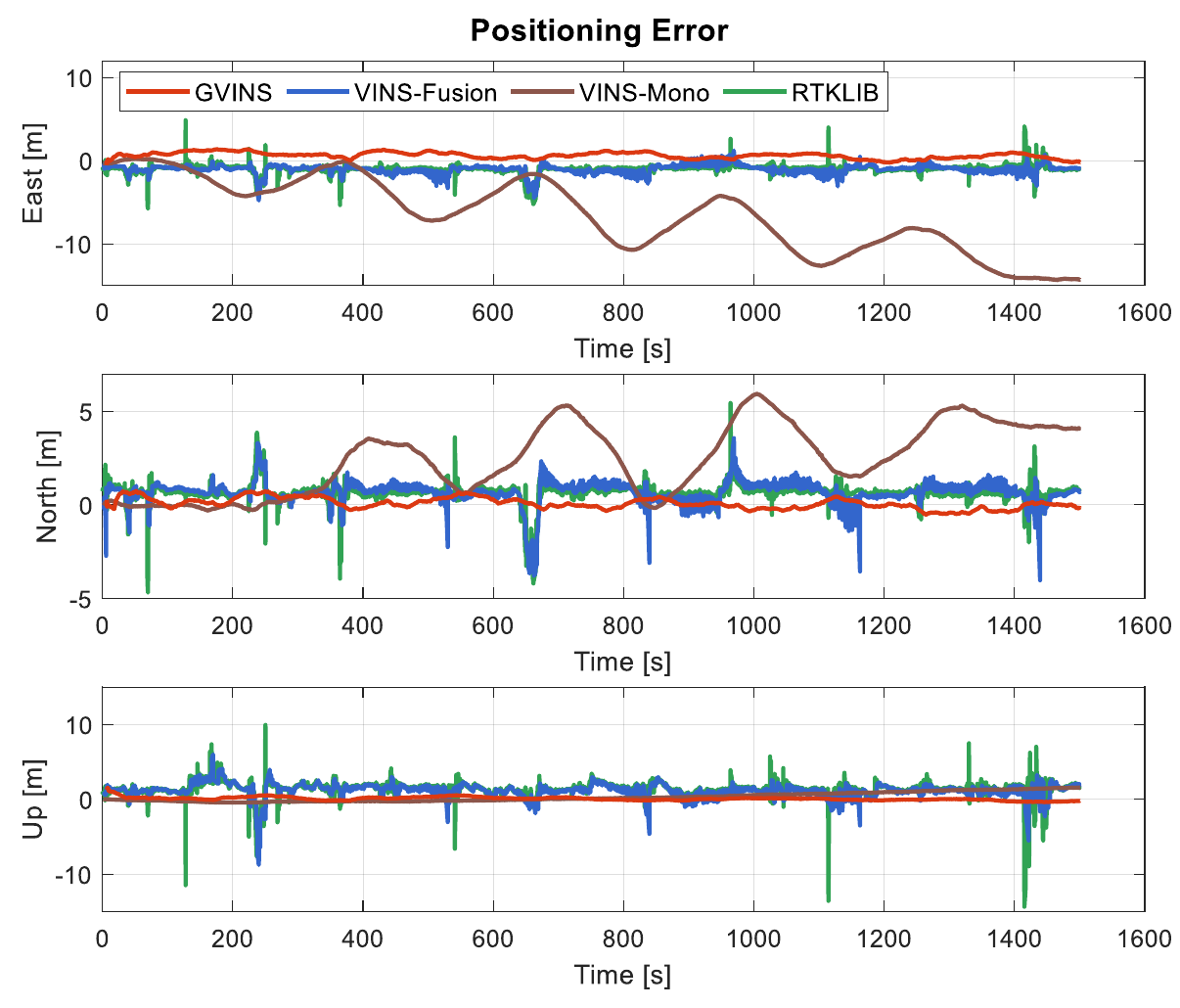}
    \caption{
        \label{fig:playground_enu_error} 
        Positioning error of GVINS, VINS-Fusion, VINS-Mono and RTKLIB at the sports field experiment. The three sub-figures correspond to the three directions of ENU frame. The result from GVINS, VINS-Fusion and RTKLIB are compared directly against the RTK ground truth without any alignment, while the result from VINS-Mono is aligned to the ground truth trajectory beforehand.}
\end{figure}

\begin{figure}
    \centering
	\vspace{0.1cm}
    \includegraphics[trim=0 0 0 0,clip,width=0.95\columnwidth]{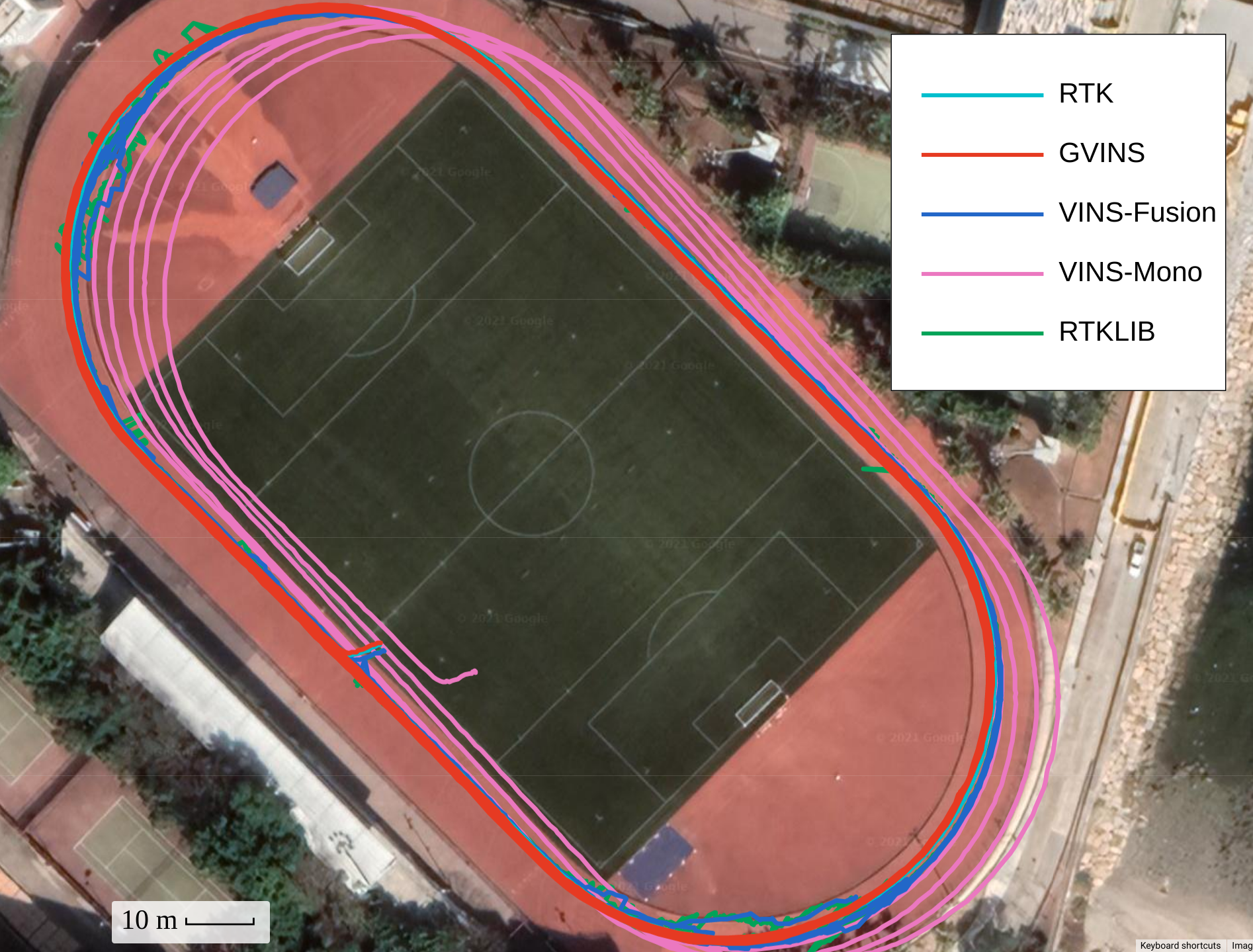}
    \caption{
        \label{fig:map_playground} 
        The trajectory of RTK, GVINS, VINS-Fusion, VINS-Mono and RTKLIB in the sports field experiment. The resulting trajectory of our proposed system is smooth and aligns well with that of the RTK. }
\end{figure}

\subsubsection{Sports Field Experiment}
\label{sssec:sports_field_experiments}
This experiment is conducted on a sports field at our campus where we follow an athletic track for 5 laps. The sports field is a typical outdoor environment with an opened area on one side and some buildings the other side. During the experiment most of the satellites are well locked and the status of RTK remains fixed throughout the whole path. In this experiment the global consistency of our estimator is examined against the repeated trajectory and the unstable signal near buildings also poses challenges to the local smoothness of the result. 

\begin{figure}
    \centering
    \includegraphics[width=0.95\columnwidth]{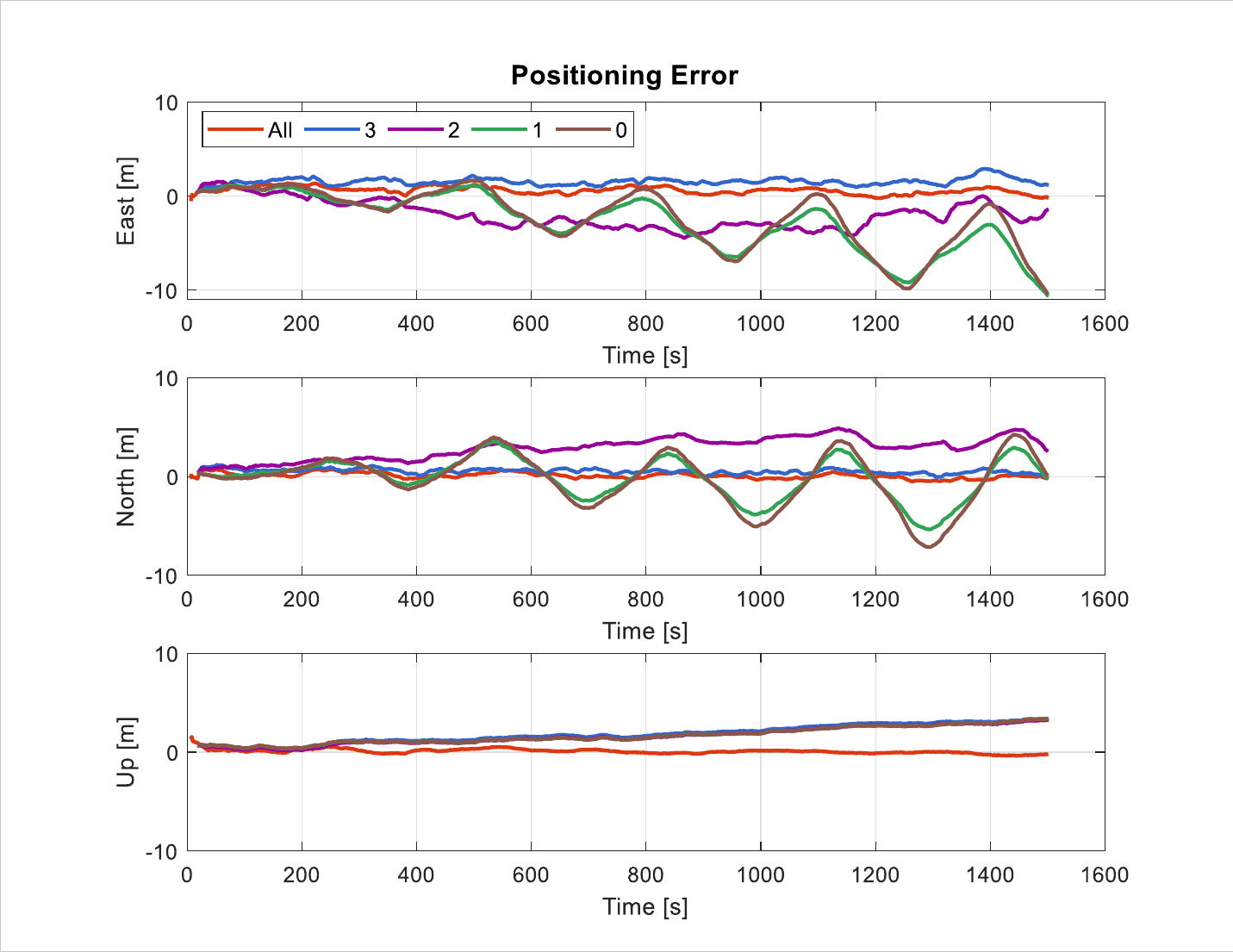}
    \caption{
        \label{fig:less_satellites_enu_error} 
		Positioning error of our proposed system in situations where the number of locked satellites is insufficient. In the ``All" setting, the system utilizes all available (around 20) satellites to perform estimation. The ``3", ``2", ``1" corresponds to cases where only that number of satellites are used in the system. When the number becomes 0, our system does not use any satellite and degrades to a VIO.}
\end{figure}

In this experiment, the GNSS-VI is initialized in $4.1\,s$ after the visual-inertial finish its alignment. The positioning error of this experiment is plotted against ENU axes as depicted in Fig.~\ref{fig:playground_enu_error}. A reference point, which is used to transform the ECEF result to a ENU frame, is arbitrarily selected on the sports field. Since VINS-Fusion, RTKLIB and our system can directly output estimation results in ECEF frame, we do not apply any alignment for their trajectories. For VINS-Mono which only gives results in local frame, we perform a 4-DOF alignment between its trajectory and the ENU path of RTK using the first 2000 poses. Note that the global positioning results from VINS-Fusion, RTKLIB and our system suffer from a certain bias due to satellites' orbit error, inaccurate atmospheric delay modeling and multipath effect, while that of VINS-Mono does not have this issue because of the pre-alignment we made.

From Fig.~\ref{fig:playground_enu_error} we see that VINS-Mono suffers from drifting among all three directions. In addition, the periodic fluctuations on horizontal directions (east and north) implies an obvious drift on the yaw estimation. On the other hand, the SPP solution from RTKLIB does not drift at all, but is highly affected by the noisy GNSS measurement. The error of VINS-Fusion is bounded as a result of combining the global information from SPP result. However, the local accuracy oscillates a lot and the local smoothness is ruined in the meantime. As a comparison, the positioning error of our proposed system does not grow with the traveled distance and is always maintained at a low level. Meanwhile, the error varies slowly and continuously, which also indicates our system effectively suppresses the noise from unstable GNSS signals. Table.~\ref{tab:rmse_table} lists the RMSE of each method and Fig.~\ref{fig:map_playground} shows final trajectories on Google Maps. The resulting 5 laps of our system overlap with each other and align well with those of RTK. Through this experiment, we show that our system is able to achieve global consistency to eliminate drifts of VIO and also preserve the local smoothness under noisy GNSS conditions.

\begin{figure}
    \centering
    % \vspace{0.5cm}
    \includegraphics[trim=0 0 0 0,clip,width=0.95\columnwidth]{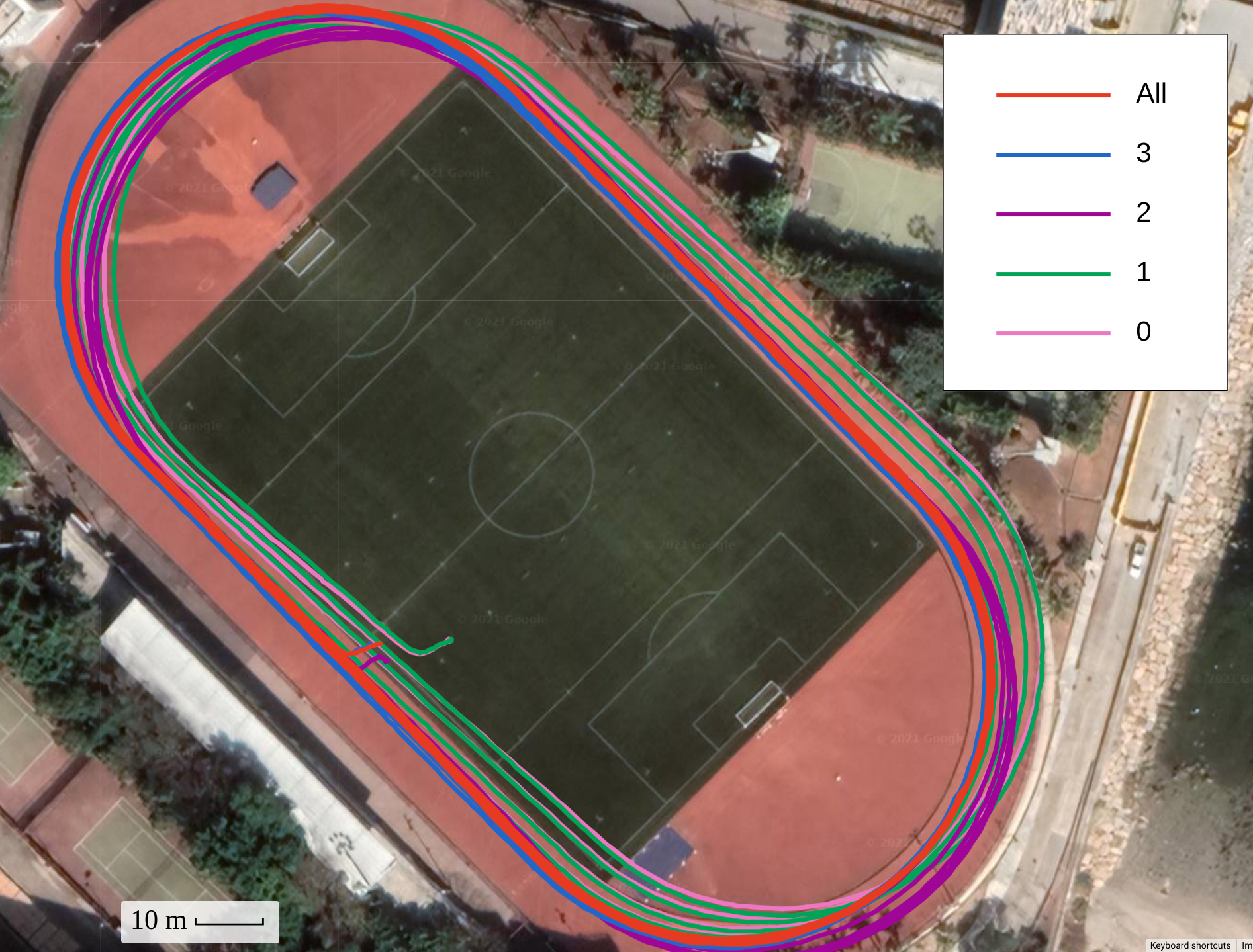}
    \caption{
        \label{fig:map_less_satellite} 
        The trajectories of our proposed system with different satellite configurations. GVINS performs best by utilizing all available satellites (``All"), and degrades to VIO with zero satellite configuration (``0"). A small bias occurs when only 3 satellites are used (``3"), and translational drift emerges when the satellite number is further reduced to 2 (``2"). If there is only 1 satellite available (``1"), yaw estimation starts to drift as well, but with a smaller magnitude compared to VIO (0 satellite).}
	\vspace{-0.5cm}
\end{figure}

\subsubsection{Insufficient Satellites Experiment}
Based on the data sequence of sports field experiment, we further investigated the degenerate case where the number of tracked satellites is less than 4. Normally there are about 20 satellites being locked in this sequence, and we intentionally remove most of the satellites in the non-linear optimization phase in order to test the system behavior. Starting from the zero-satellite setting, we sequentially add satellite G2, G13 and G5, which are well tracked during the experiment, to the system to simulate the one, two, and three-satellite situations respectively. In this experiment we only use satellites from a single constellation (GPS) because the general case where $M$ satellites coming from $N$ constellations is equivalent to the $(M-N+1)$ single-constellation case due to unknown clock offsets between different systems. It is worth to mention that our system naturally degrades to a VIO when there is no satellite available.

The positioning error with 5 different settings is illustrated in Fig.~\ref{fig:less_satellites_enu_error}. Obviously our system performs best in the normal setting where all available satellites are used for estimation. In the up direction, the errors of all other 4 configurations accumulate in a similar manner. This indicts that the drift in the up direction can no longer be eliminated with 3 satellites or less. In terms of horizontal directions, no accumulated error but only a small bias occurs for the three-satellite setting, which means our system is still able to suppress drifts in east, north and yaw directions. If the number of satellites is further reduced to 2, the horizontal positioning error starts growing with the traveled distance, and we observed small periodic fluctuations in north direction which coincides with that of VIO. This implies that the drift in horizontal plane occurs and yaw error also emerges although the magnitude is very small. Finally with the one-satellite configuration, accumulated errors occur on all four unobservable directions of VIO. However, the error of the yaw component is still smaller compared to that of VIO, which can be inferred from the amplitude of the sine-wave-like error curve. The final trajectories with different satellite settings is shown in Fig.~\ref{fig:map_less_satellite}. Through this experiment, we claim that our system gradually degrades to different extents when the number of locked satellites varies from 3 to 0. However, the proposed system outperforms a pure VIO in all different settings which indicts that our tightly fusion approach can still gain information from limited satellites.

\subsubsection{Indoor-outdoor Experiment}
The GNSS-VI initialization takes $9.0\,s$ in this experiment, with the majority of time waiting for GNSS navigation messages. This experiment, through which we aim to test the robustness of our system, is performed in a complex indoor-outdoor environment. The path of this experiment goes through many challenging scenarios which may bring a single-sensor-based system to failure. For example, no features are detected and tracked in dim or bright area, and the GNSS signal is highly corrupted or totally unavailable in cluttered or indoor environment. In addition, the path is similar to the one in a typical exploration task where no large loops exist, thus drifting is inevitable for any visual-inertial SLAM system. The overall distance of the resulting trajectory is over 3 kilometers and the altitude change is around 130 meters. 

\begin{figure}
    \centering
    \includegraphics[width=0.95\columnwidth]{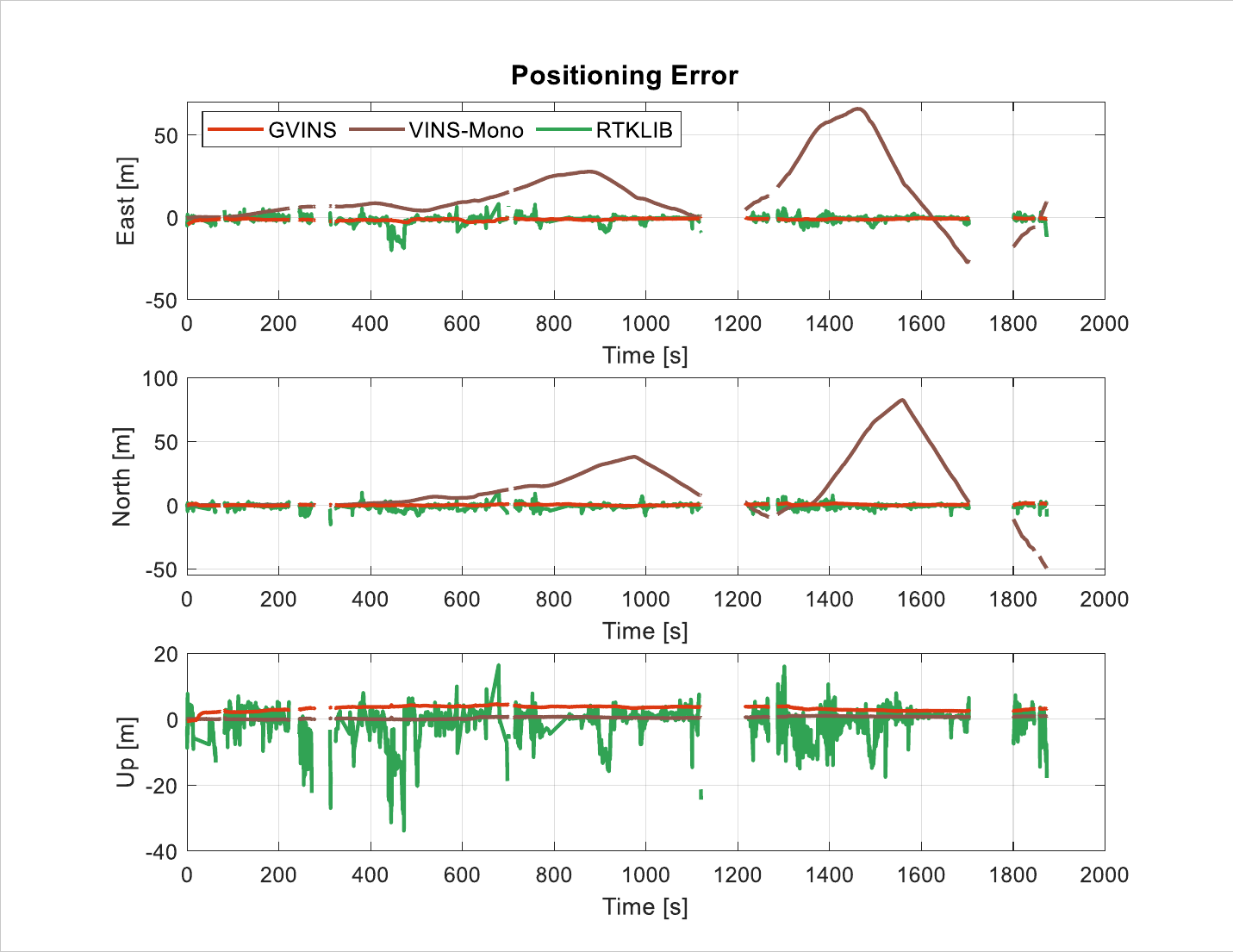}
    \caption{
        \label{fig:longpath_enu_error} 
        Positioning error of GVINS, VINS-Mono and RTKLIB in the complex indoor-outdoor experiment. We only compare with the RTK fix solutions, so the gaps in the figure corresponding to the situations where ground truth is not available. The result of VINS-Fusion is not shown because of huge errors and oscillations.}
	\vspace{-0.5cm}
\end{figure}

\begin{figure}
    \centering
    \includegraphics[width=0.95\columnwidth]{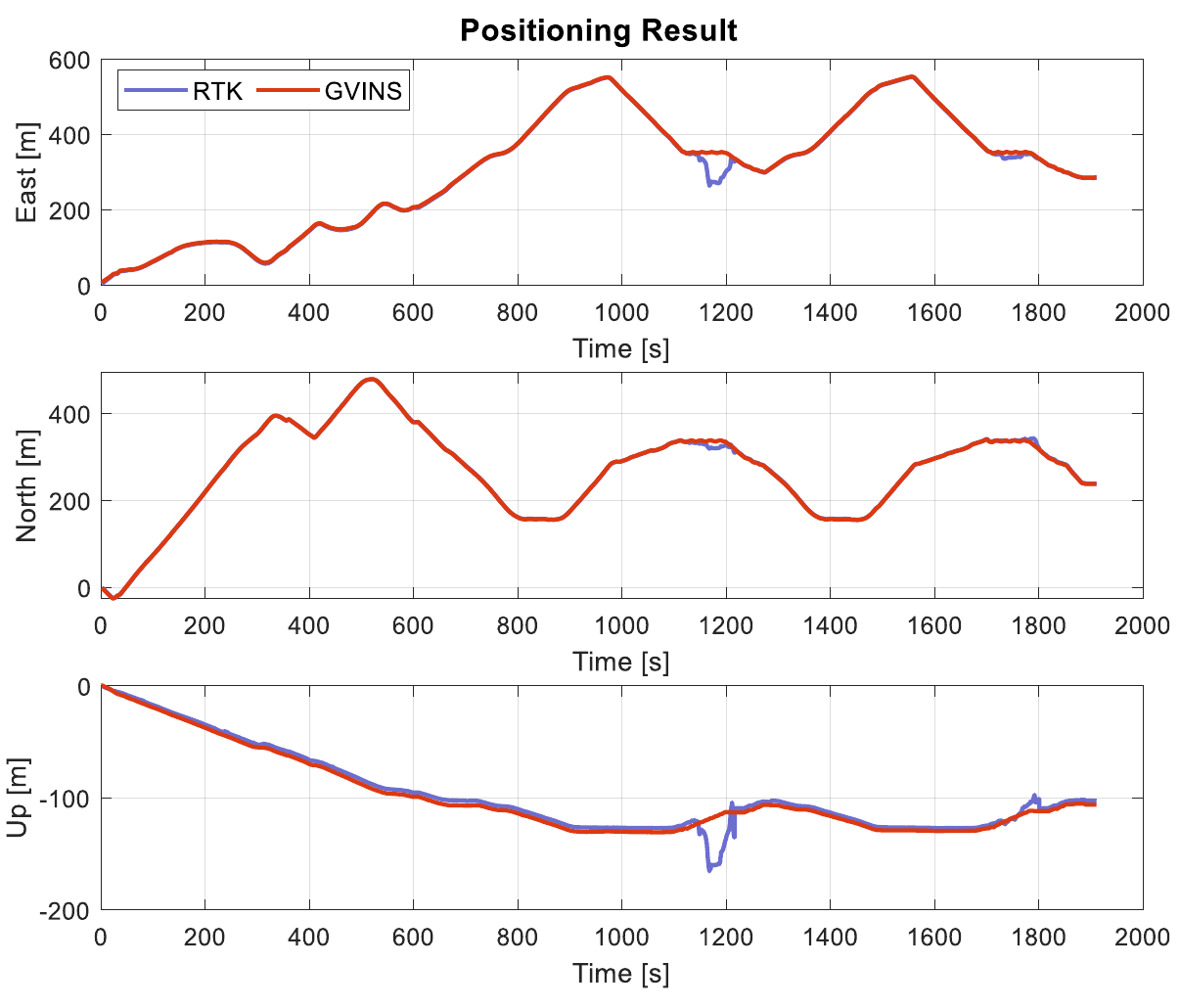}
    \caption{
        \label{fig:longpath_enu_result} 
        Positioning result of RTK and GVINS in the complex indoor-outdoor experiment. }
\end{figure}

\begin{figure}
    \centering
    \includegraphics[width=0.95\columnwidth]{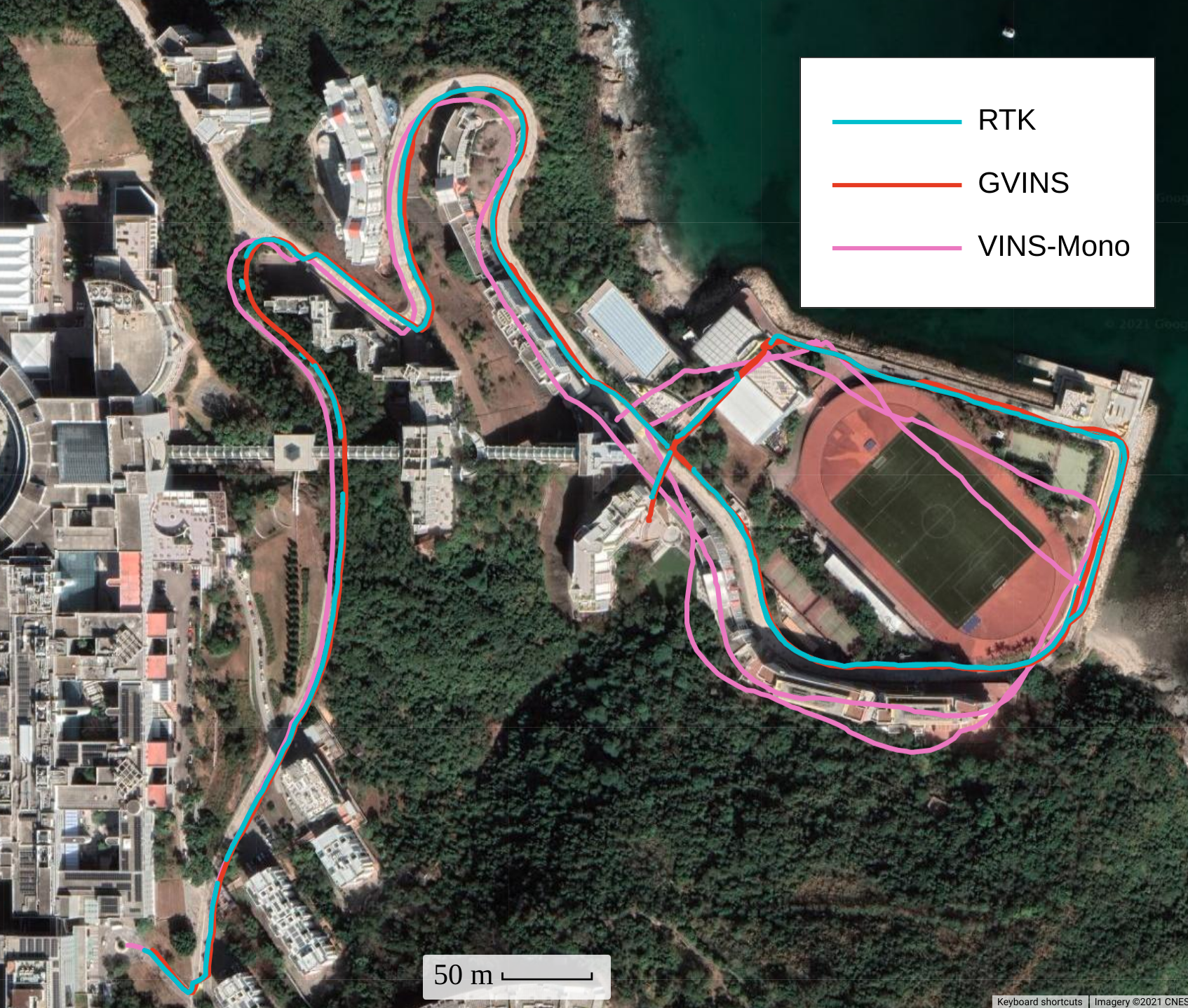}
    \caption{
        \label{fig:long_path_trajectory} 
        Final trajectories in the complex indoor-outdoor experiment. The result of RTKLIB and VINS-Fusion are not plotted because of large noise and jitters. The discontinuities on the RTK path is the result of bad GNSS signal and fix-lost events.}
	\vspace{-0.5cm}
\end{figure}

Fig.~\ref{fig:longpath_enu_error} shows the ENU positioning error on the indoor-outdoor sequence. During this experiment the RTK ground truth is no longer always available because of the GNSS-unfriendly environment. Thus we only compare with segments where RTK is in fix status. The gaps around 300s occurs when we were under a bridge and passing through the woods which blocked most of the sky, and the blanks around 1200s and 1800s correspond to the situation where we were going up the indoor stairs. The result of VINS-Fusion is not shown in the figure because of huge errors and oscillations. It can be observed from the figure that VINS-Mono still experiences large accumulated errors on horizontal and yaw directions, while the error in the up direction is smaller than the previous experiment because of the attitude excitation on this sequence. The result of RTKLIB, although does not drift, varies a lot around the ground truth value. Those oscillations indict the condition of GNSS signal and severely affect the performance of VINS-Fusion. Our proposed system outperforms other three approaches in terms of positioning error and overcomes the harsh condition brought by the noisy GNSS measurement. The result of our system still has a bias on the up direction because of imperfect GNSS modelling and various error sources, while the up error of VINS-Mono starts from zero because of pre-alignment. The final trajectories of RTK, aligned VINS-Mono and our system is shown in Fig.~\ref{fig:long_path_trajectory}. The figure shows that both VINS-Mono and our proposed system work well across the whole sequence, although obvious drift occurs on the result of VINS-Mono. The discontinuities on the trajectory of RTK is the result of cluttered and indoor environment. The trajectory of our system follows the RTK result well, and the positioning result, even in GNSS-unfriendly area, can be effectively recovered. Although the duration where RTK fails is short in the whole sequence, the impact can be significant. As shown in Fig.~\ref{fig:longpath_enu_result}, the result of RTK is smooth and aligns well with that of GVINS when GNSS is reliable. However, the solution reported by RTK results in an error of up to 80 meters during GNSS outage, and such behavior is catastrophic for any location-based tasks. The final RMSE of all four approaches is shown in table.~\ref{tab:rmse_table}. 

\subsubsection{GNSS Factor Experiment}
\begin{figure}
	\centering
    \includegraphics[width=0.95\columnwidth]{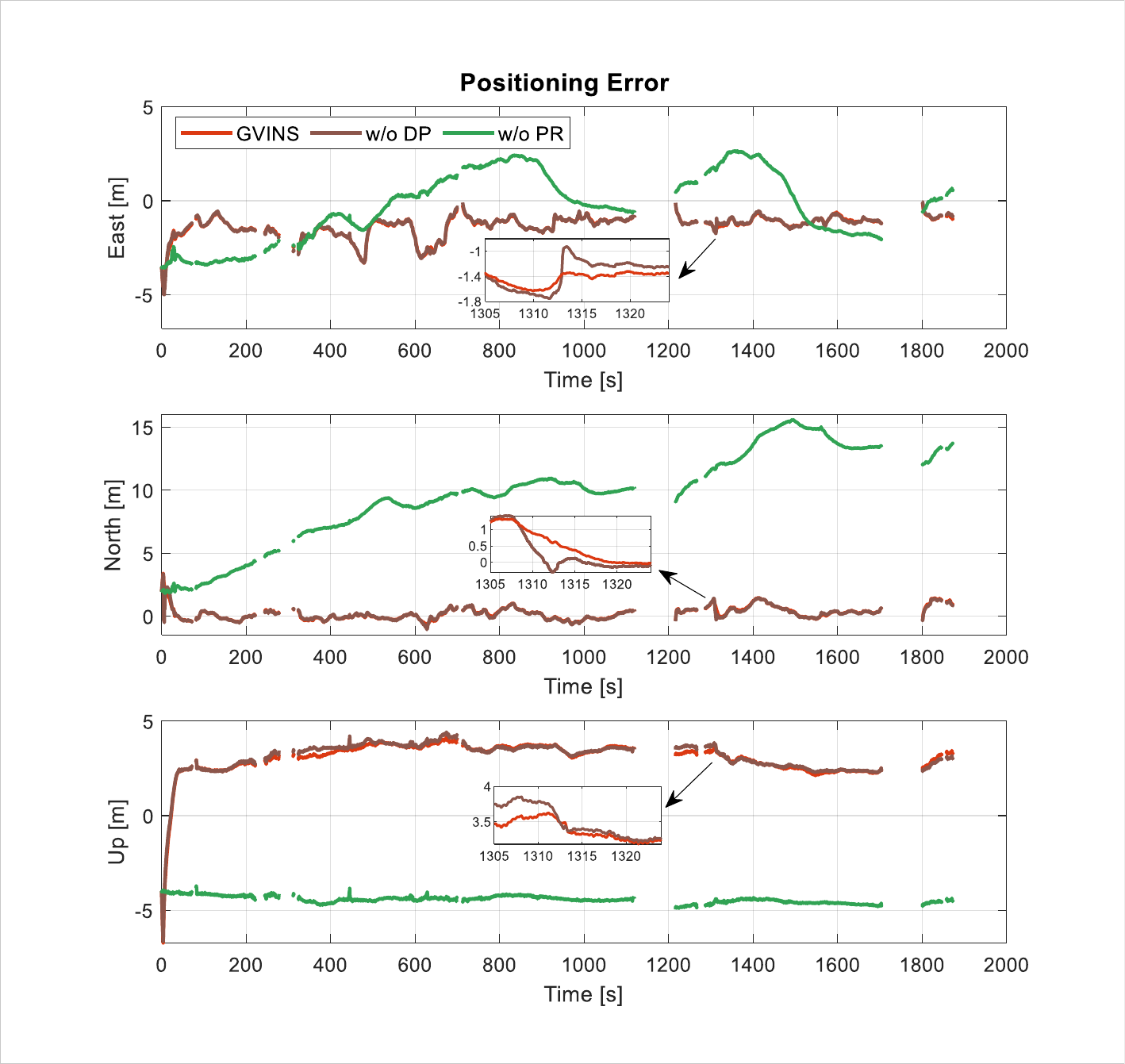}
    \caption{
        \label{fig:gnss_factor_enu_error} 
        Positioning error of normal GVINS, GVINS w/o Doppler factor and GVINS w/o code pseudorange factor.}
\end{figure}

Based on the previous indoor-outdoor sequence, we further investigate the role of each GNSS measurement(i.e. code pseudorange, Doppler shift) on the performance of our proposed system. By removing the corresponding graph factor after initialization phase, we obtain the positioning error on code pseudorange-only and Doppler-only configurations as depicted in Fig.~\ref{fig:gnss_factor_enu_error}. In the situation where we only employ Doppler shift measurement, an obvious drift occurs as the system no longer has global position constraints. In addition, the initialization error, which is inevitable because we initialize from only a short window of measurements, cannot be eliminated and acts like a bias subsequently. If we instead conduct the code pseudorange-only optimization, the system behaves like a normal GVINS, e.g. the system does not drift any more and the initialization error can be eliminated after a short period. However, as the code pseudorange measurement tends to be noisy and receiver clock biases are no longer constrained by Doppler shift, the smoothness of the estimation result is affected by the unstable signal, as shown in the magnified portion of Fig.~\ref{fig:gnss_factor_enu_error}. Through this experiment, we show that the code pseudorange measurement is the key to eliminating the accumulated drift of VIO. However, with the aid of the Doppler shift measurement, the estimation result tends to be smoother under unstable GNSS conditions.

\subsubsection{Urban Driving Experiment}
\begin{figure}[ht]
	\centering
	\subfigure[Urban canyon]{%
	\includegraphics[scale=0.15]{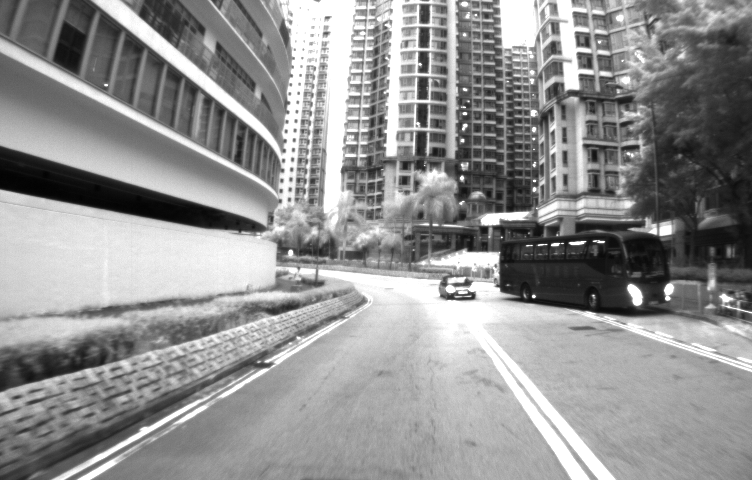}
	\label{fig:urban_canyon}}
	\quad
	\subfigure[Dynamic and dark scene]{%
	\includegraphics[scale=0.15]{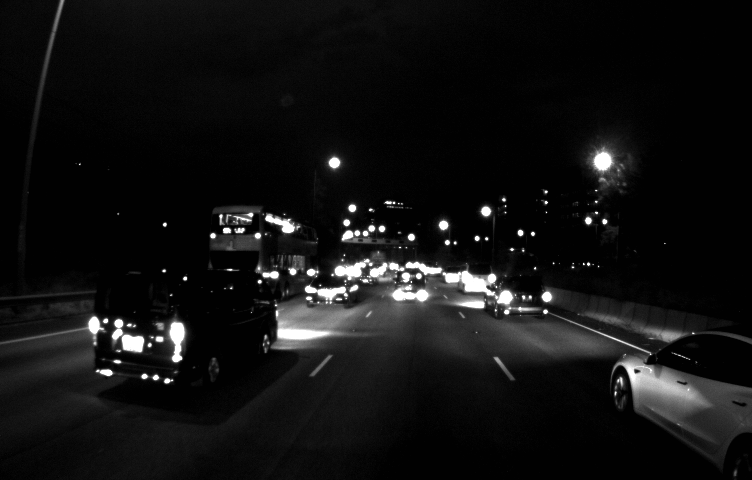}
	\label{fig:dark_env}}
	\caption{Two image samples illustrate the challenging situations in the urban driving experiment. In the left image the GNSS receiver is surrounded by high-rise buildings, where multipath effect is obvious. The right image shows a highly dynamic scenario with low illumination and a high traffic flow on an expressway.}
	\label{fig:urban_image_samples}
\end{figure}

In this experiment we test our system with a challenging urban driving scenario in one of the most populous districts of Hong Kong. The experiment begins at dusk and lasts over 40 minutes until complete dark, with a total distance of 22.9 km. The data sequence covers heterogeneous situations, such as day and night, urban canyon and open-sky outdoors, etc. The challenging cases, including high-rise buildings, low illumination, fast movement and highly dynamic environments, make it impracticable for a single-sensor based algorithm to deal with. Two image samples from the data sequence are shown in Fig.~\ref{fig:urban_image_samples}.

During the experiment, the GNSS outage occur constantly even in outdoor environments, because of the traffic signs and bridges above the road. In addition, severe multipath effect is observed on the GNSS measurements when the receiver is surrounded by high-rise buildings in urban canyon. To this end, a robust norm is applied on the code pseudorange and Doppler shift factors to re-weight GNSS outliers. 

\begin{figure}
    \centering
    \includegraphics[width=0.95\columnwidth]{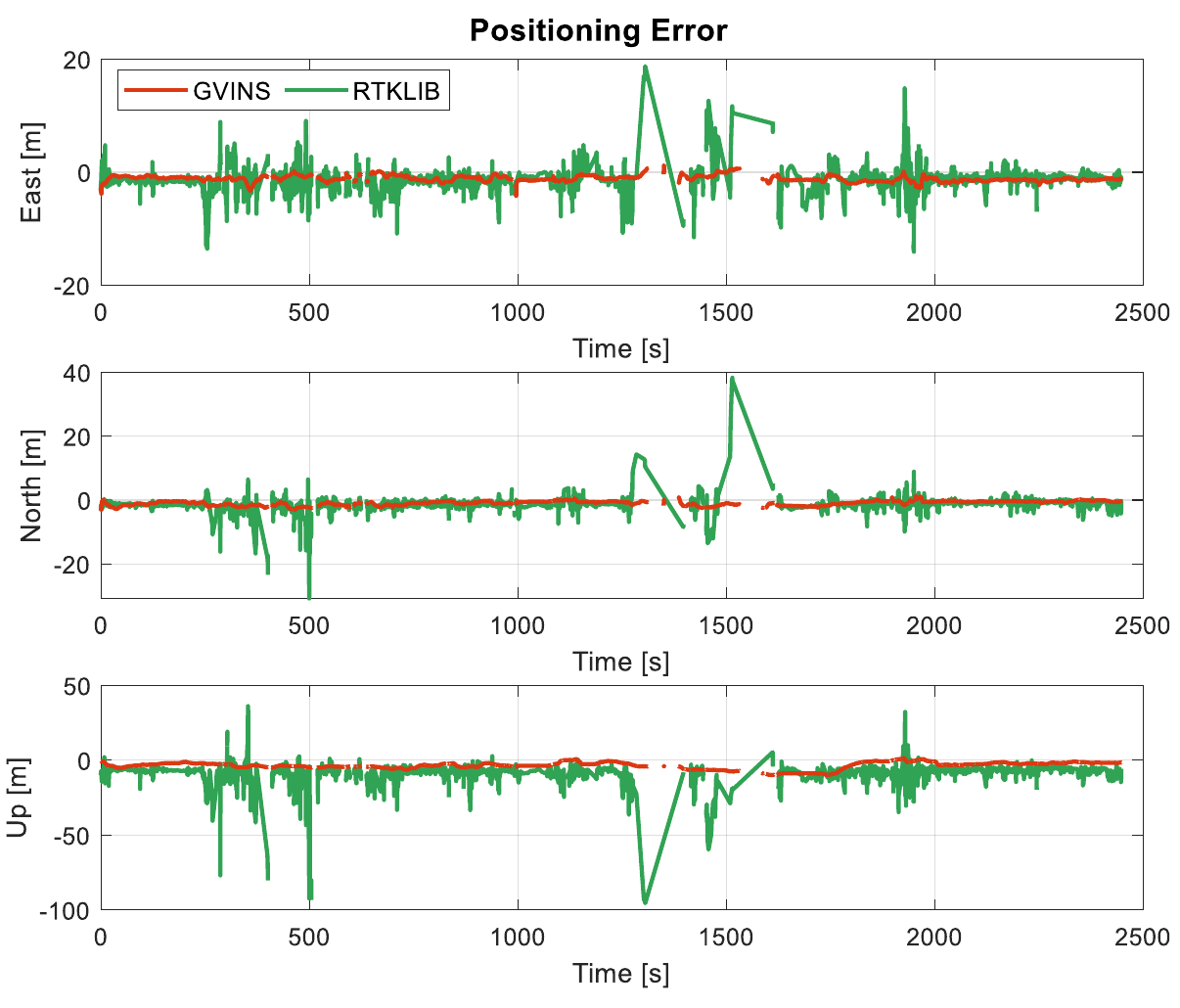}
    \caption{
        \label{fig:urban_driving_enu_error} 
        Positioning error of GVINS, VINS-Mono and RTKLIB in the urban driving experiment. The gaps in the figure corresponding to the RTK non-fixed status. The results of VINS-Fusion and VINS-Mono are not shown because of failure.}
	\vspace{-0.5cm}
\end{figure}

On this sequence, the VINS-Mono, which only uses visual and inertial sensors to perform estimation, fails at 1200s when the sky becomes dark and many vehicles pass by. The failure of VINS-Mono occurs at $54 \%$ of the total distance, with a RMSE of $760.22 ~m$ indicting a large drift. The loosely-coupled GNSS-visual-inertial algorithm, VINS-Fusion, does not explicitly report any failure. However, hugh oscillations are observed on its result, with the corresponding RMSE at the order of $10^5 ~m$. To this end, we also mark the result of VINS-Fusion as a failure case. 

In this experiment, the GNSS-VI is successfully initialized in $2.0\,s$ after the visual-inertial initialization finishes. Fig.~\ref{fig:urban_driving_enu_error} shows the positioning error of GVINS and RTKLIB on three axes of ENU frame respectively. The extreme errors from the result of RTKLIB, which we define as above $100\,m$, are not shown to limit the scale of the plot. The large-magnitude oscillations of RTKLIB on this data sequence clearly illustrate the terrible quality of GNSS signal in the harsh environment, especially around $400\,s$ and $1350\,s$, where the receiver is surrounded by high-rise buildings and multipath effect is severe. Our proposed system, GVINS, survives through the whole sequence, which again proves the availability and robustness of our system. The slowly varying and well bounded positioning error of GVINS in Fig.~\ref{fig:urban_driving_enu_error} shows the local smoothness and global consistency properties of the proposed method. 

\begin{figure}
    \centering
    \includegraphics[width=0.95\columnwidth]{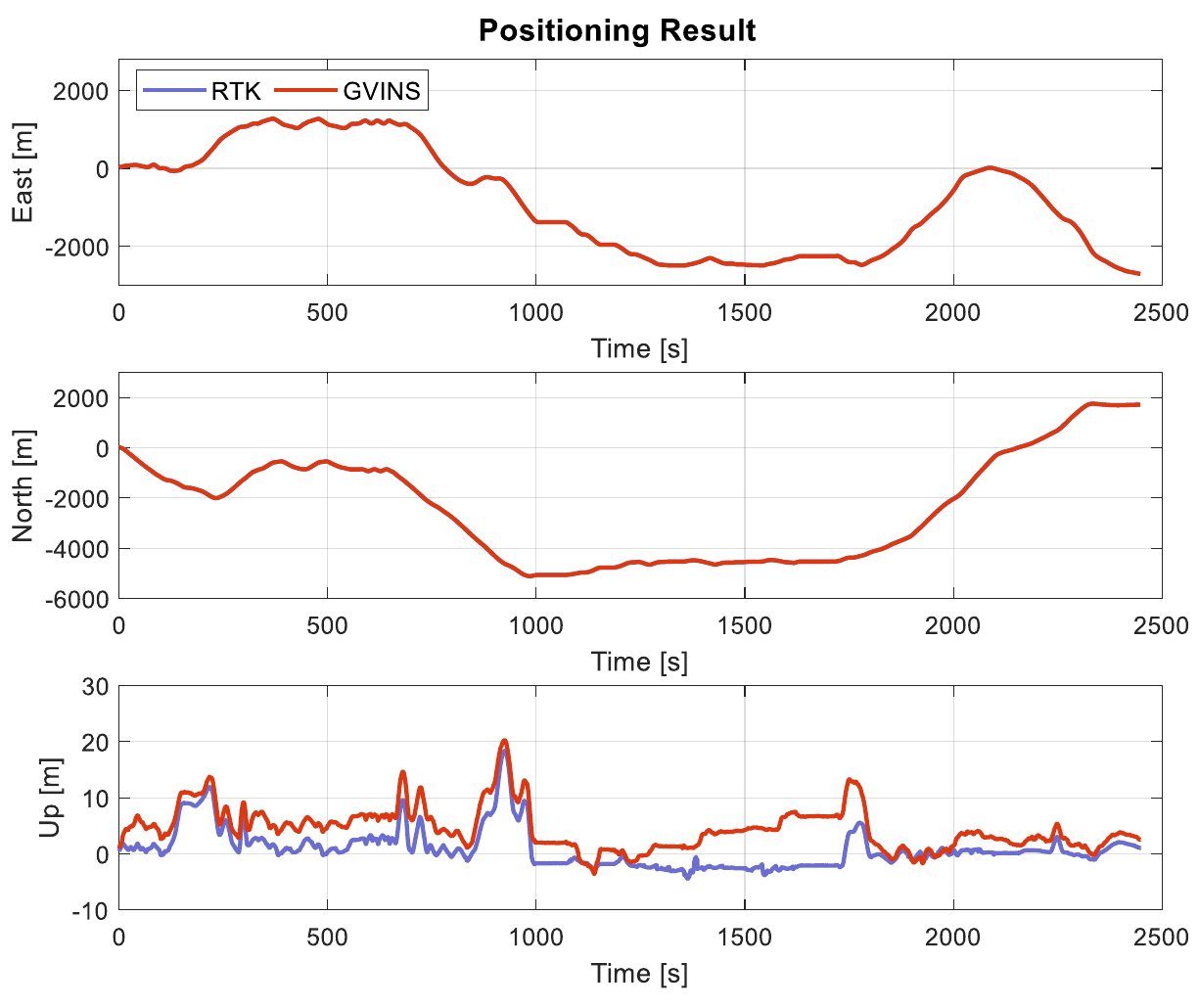}
    \caption{
        \label{fig:urban_driving_enu_result} 
        Positioning result of RTK and GVINS in the challenging urban driving experiment. }
    \vspace{-0.6cm}
\end{figure}

\begin{figure}
    \centering
	\vspace{0.1cm}
    \includegraphics[width=0.95\columnwidth]{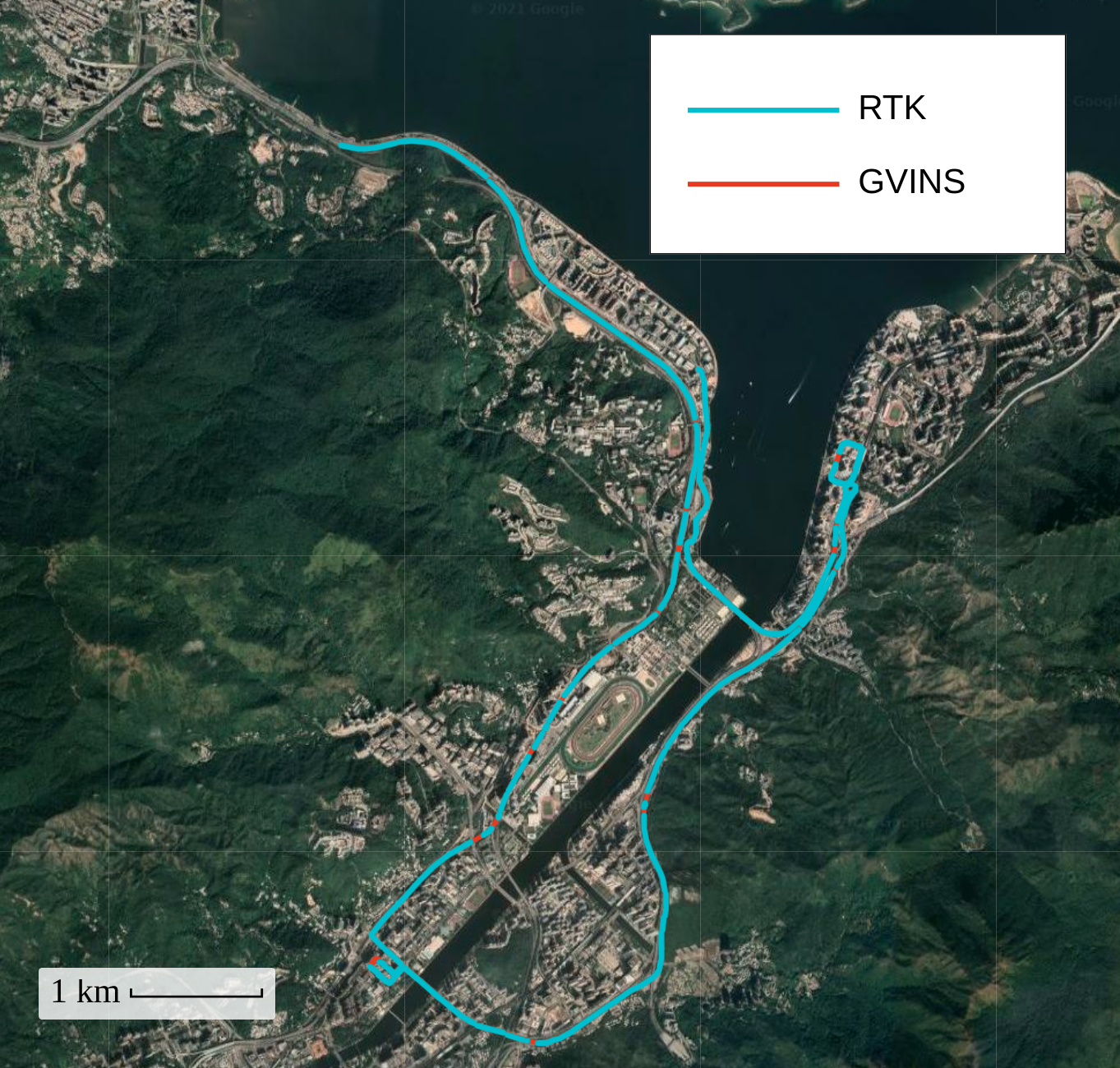}
    \caption{
        \label{fig:map_urban_driving} 
        The trajectories of RTK and GVINS in the urban driving experiment. The two paths totally align with each other. The trajectory of RTK is plotted on the top of that of GVINS so that RTK non-fixed status can be clearly shown by the discontinuities.}
\end{figure}

Fig.~\ref{fig:urban_driving_enu_result} illustrates the positioning results of RTK and GVINS. From Fig.~\ref{fig:urban_driving_enu_result} we see that the trajectory of our system aligns well with that of RTK on horizontal directions. Since we do not perform any alignment on the result of GVINS, a obvious bias, in addition to the varying error, can be observed on the vertical direction. The RMSE of GVINS and RTKLIB is also included in Table.~\ref{tab:rmse_table}, and the fields of VINS-Mono and VINS-Fusion are marked as N/A because of failure. The final trajectories on Google Maps are depicted in Fig.~\ref{fig:map_urban_driving}, where the result of RTK is plotted on the top of GVINS. Note that the GNSS outage occurs constantly even on the open-sky expressway because of the traffic signs and viaducts. By stack the trajectory of RTK on the top of that of GVINS, the discontinuities on the path of RTK, corresponding to the RTK non-fixed status over long distance, can be clearly illustrated in the figure. Due to the large scale of the map, the frequent short-term RTK outages cannot be observed. 

In terms of the computational time, the feature detection and tracking, which are same for VINS-Mono, VINS-Fusion and GVINS, costs $7.28 ~ms$ per frame. The window optimization of VINS-Mono takes $21.76 ~ms$ on average. The time spent in the pose graph module of VINS-Fusion, which is used to fuse GNSS solution with visual-inertial odometry, grows as the travelled distance increases. The lower limit is $1.12 ~ms$ at the beginning and the upper bound is $1018.46 ~ms$ in the end, with an average value of $404.83 ~ms$. In contrast, our proposed GVINS only needs $21.91 ~ms$ on the window optimization thanks to the tightly-coupled and sliding-window approaches we adopted. Considering the 20-Hz camera we used in our experiments, our system can safely run at real-time while obvious lags may be observed in the case of VINS-Fusion as the travelled distance grows.

\section{Conclusion}
\label{sec:conclusion}
In this paper, we propose a tightly-coupled system to fuse measurements from camera, IMU and GNSS receiver under a non-linear optimization-based framework. Our system starts with an initialization phase, during which a coarse-to-fine procedure is employed to online calibrate the transformation between the local and global frames. In the optimization phase, GNSS raw measurements are modelled and formulated using the probabilistic factor graph. The degenerate cases are considered and carefully handled to keep the system robust in the complex environment. We conduct experiments in both simulation and real-world environments to evaluate the performance of our system, and the results show that our system effectively eliminates the accumulated drift and preserves the local accuracy of a typical VIO system. To this end, we state that our system can achieve both local smoothness and global consistency. 

In future work, the theoretical observability analysis will be conducted under various degenerate situations and we aim to build an online observability-aware state estimator to deal with complex environments and possible sensor failures. In addition, we are also interested in reducing the absolute positioning error by GNSS measurements combination\cite{blewitt1990automatic} or \ac{PPP} \cite{zumberge1997precise} techniques to handle distributed localization tasks in swarm systems.

\ifCLASSOPTIONcaptionsoff
  \newpage
\fi

\bibliography{GVINS_arxiv_v3.bib}

\begin{thebibliography}{10}
\providecommand{\url}[1]{#1}
\csname url@samestyle\endcsname
\providecommand{\newblock}{\relax}
\providecommand{\bibinfo}[2]{#2}
\providecommand{\BIBentrySTDinterwordspacing}{\spaceskip=0pt\relax}
\providecommand{\BIBentryALTinterwordstretchfactor}{4}
\providecommand{\BIBentryALTinterwordspacing}{\spaceskip=\fontdimen2\font plus
\BIBentryALTinterwordstretchfactor\fontdimen3\font minus
  \fontdimen4\font\relax}
\providecommand{\BIBforeignlanguage}[2]{{%
\expandafter\ifx\csname l@#1\endcsname\relax
\typeout{** WARNING: IEEEtran.bst: No hyphenation pattern has been}%
\typeout{** loaded for the language `#1'. Using the pattern for}%
\typeout{** the default language instead.}%
\else
\language=\csname l@#1\endcsname
\fi
#2}}
\providecommand{\BIBdecl}{\relax}
\BIBdecl

\bibitem{huang2014towards}
G.~Huang, M.~Kaess, and J.~J. Leonard, ``Towards consistent visual-inertial
  navigation,'' in \emph{Proc. of the {IEEE} Int. Conf. on Robot. and Autom.},
  2014, pp. 4926--4933.

\bibitem{MouRou0704}
A.~I. Mourikis and S.~I. Roumeliotis, ``A multi-state constraint {K}alman
  filter for vision-aided inertial navigation,'' in \emph{Proc. of the {IEEE}
  Int. Conf. on Robot. and Autom.}, Roma, Italy, Apr. 2007, pp. 3565--3572.

\bibitem{LiMou1305}
M.~Li and A.~Mourikis, ``High-precision, consistent {EKF}-based visual-inertial
  odometry,'' \emph{Int. J. Robot. Research}, vol.~32, no.~6, pp. 690--711, May
  2013.

\bibitem{wu2015square}
K.~Wu, A.~Ahmed, G.~A. Georgiou, and S.~I. Roumeliotis, ``A square root inverse
  filter for efficient vision-aided inertial navigation on mobile devices.'' in
  \emph{Robot.: Science and Systems}, 2015, p.~2.

\bibitem{LeuFurRab1306}
S.~Leutenegger, S.~Lynen, M.~Bosse, R.~Siegwart, and P.~Furgale,
  ``Keyframe-based visual-inertial odometry using nonlinear optimization,''
  \emph{Int. J. Robot. Research}, vol.~34, no.~3, pp. 314--334, Mar. 2014.

\bibitem{qin2018vins}
T.~Qin, P.~Li, and S.~Shen, ``Vins-mono: A robust and versatile monocular
  visual-inertial state estimator,'' \emph{{IEEE} Trans. Robot.}, vol.~34,
  no.~4, pp. 1004--1020, 2018.

\bibitem{6289875}
C.~V. Angelino, V.~R. Baraniello, and L.~Cicala, ``Uav position and attitude
  estimation using imu, gnss and camera,'' in \emph{Proc. of the Int. Conf. on
  Information Fusion}, 2012, pp. 735--742.

\bibitem{lynen2013robust}
S.~Lynen, M.~W. Achtelik, S.~Weiss, M.~Chli, and R.~Siegwart, ``A robust and
  modular multi-sensor fusion approach applied to mav navigation,'' in
  \emph{Proc. of the {IEEE/RSJ} Int. Conf. on Intell. Robots and Syst.}\hskip
  1em plus 0.5em minus 0.4em\relax IEEE, 2013, pp. 3923--3929.

\bibitem{SheMulMic1405}
S.~Shen, Y.~Mulgaonkar, N.~Michael, and V.~Kumar, ``Multi-sensor fusion for
  robust autonomous flight in indoor and outdoor environments with a rotorcraft
  {MAV},'' in \emph{Proc. of the {IEEE} Int. Conf. on Robot. and Autom.}, Hong
  Kong, China, May 2014, pp. 4974--4981.

\bibitem{mascaro2018gomsf}
R.~Mascaro, L.~Teixeira, T.~Hinzmann, R.~Siegwart, and M.~Chli, ``Gomsf:
  Graph-optimization based multi-sensor fusion for robust uav pose
  estimation,'' in \emph{Proc. of the {IEEE} Int. Conf. on Robot. and Autom.},
  2018, pp. 1421--1428.

\bibitem{8968519}
Y.~Yu, W.~Gao, C.~Liu, S.~Shen, and M.~Liu, ``A gps-aided omnidirectional
  visual-inertial state estimator in ubiquitous environments,'' in \emph{Proc.
  of the {IEEE/RSJ} Int. Conf. on Intell. Robots and Syst.}, 2019, pp.
  7750--7755.

\bibitem{qin2019general}
\BIBentryALTinterwordspacing
T.~Qin, S.~Cao, J.~Pan, and S.~Shen, ``A general optimization-based framework
  for global pose estimation with multiple sensors,'' \emph{CoRR}, vol.
  abs/1901.03642, 2019. [Online]. Available:
  \url{http://arxiv.org/abs/1901.03642}
\BIBentrySTDinterwordspacing

\bibitem{li2021semi}
X.~Li, X.~Wang, J.~Liao, X.~Li, S.~Li, and H.~Lyu, ``Semi-tightly coupled
  integration of multi-gnss ppp and s-vins for precise positioning in
  gnss-challenged environments,'' \emph{Satellite Navigation}, vol.~2, no.~1,
  pp. 1--14, 2021.

\bibitem{gakne2018tightly}
P.~V. Gakne and K.~O’Keefe, ``Tightly-coupled gnss/vision using a
  sky-pointing camera for vehicle navigation in urban areas,'' \emph{Sensors},
  vol.~18, no.~4, p. 1244, 2018.

\bibitem{schreiber2016vehicle}
M.~Schreiber, H.~K{\"o}nigshof, A.-M. Hellmund, and C.~Stiller, ``Vehicle
  localization with tightly coupled gnss and visual odometry,'' in \emph{Proc.
  of the IEEE Sym. on Intell. Vehicles}, 2016, pp. 858--863.

\bibitem{6851506}
D.~P. Shepard and T.~E. Humphreys, ``High-precision globally-referenced
  position and attitude via a fusion of visual slam, carrier-phase-based gps,
  and inertial measurements,'' in \emph{Proc. of the IEEE/ION Sym. on Pos. Loc.
  and Nav.}, 2014, pp. 1309--1328.

\bibitem{li2019tight}
T.~Li, H.~Zhang, Z.~Gao, X.~Niu, and N.~El-Sheimy, ``Tight fusion of a
  monocular camera, mems-imu, and single-frequency multi-gnss rtk for precise
  navigation in gnss-challenged environments,'' \emph{Remote Sensing}, vol.~11,
  no.~6, p. 610, 2019.

\bibitem{yoder2020multi}
J.~E. Yoder, P.~A. Iannucci, L.~Narula, and T.~E. Humphreys, ``Multi-antenna
  vision-and-inertial-aided cdgnss for micro aerial vehicle pose estimation,''
  in \emph{Proc. of the Int. Tech. Meet. of the Sate. Div. of The Inst. of
  Nav.}, 2020, pp. 2281--2298.

\bibitem{5507322}
A.~Soloviev and D.~Venable, ``Integration of gps and vision measurements for
  navigation in gps-challenged environments,'' in \emph{Proc. of the IEEE/ION
  Sym. on Pos. Loc. and Nav.}, 2010, pp. 826--833.

\bibitem{won2014gnss}
D.~H. Won, E.~Lee, M.~Heo, S.~Sung, J.~Lee, and Y.~J. Lee, ``Gnss integration
  with vision-based navigation for low gnss visibility conditions,'' \emph{GPS
  solutions}, vol.~18, no.~2, pp. 177--187, 2014.

\bibitem{DBLP:journals/corr/abs-2010-11675}
\BIBentryALTinterwordspacing
J.~Liu, W.~Gao, and Z.~Hu, ``Optimization-based visual-inertial {SLAM} tightly
  coupled with raw {GNSS} measurements,'' \emph{CoRR}, vol. abs/2010.11675,
  2020. [Online]. Available: \url{https://arxiv.org/abs/2010.11675}
\BIBentrySTDinterwordspacing

\bibitem{saastamoinen1972contributions}
J.~Saastamoinen, ``Contributions to the theory of atmospheric refraction,''
  \emph{Bulletin G{\'e}od{\'e}sique (1946-1975)}, vol. 105, no.~1, pp.
  279--298, 1972.

\bibitem{klobuchar1987ionospheric}
J.~A. Klobuchar, ``Ionospheric time-delay algorithm for single-frequency gps
  users,'' \emph{IEEE Trans. on aero. and elec. syst.}, no.~3, pp. 325--331,
  1987.

\bibitem{kaplan2005understanding}
E.~Kaplan and C.~Hegarty, \emph{Understanding GPS: principles and
  applications}.\hskip 1em plus 0.5em minus 0.4em\relax Artech house, 2005.

\bibitem{SheMicKum1505}
S.~Shen, N.~Michael, and V.~Kumar, ``Tightly-coupled monocular visual-inertial
  fusion for autonomous flight of rotorcraft {MAV}s,'' in \emph{Proc. of the
  {IEEE} Int. Conf. on Robot. and Autom.}, Seattle, WA, May 2015.

\bibitem{forster2017manifold}
C.~Forster, L.~Carlone, F.~Dellaert, and D.~Scaramuzza, ``On-manifold
  preintegration for real-time visual-inertial odometry,'' \emph{{IEEE} Trans.
  Robot.}, vol.~33, no.~1, pp. 1--21, 2017.

\bibitem{7139939}
S.~Shen, N.~Michael, and V.~Kumar, ``Tightly-coupled monocular visual-inertial
  fusion for autonomous flight of rotorcraft mavs,'' in \emph{Proc. of the
  {IEEE} Int. Conf. on Robot. and Autom.}, 2015, pp. 5303--5310.

\bibitem{910572}
F.~R. Kschischang, B.~J. Frey, and H.-A. Loeliger, ``Factor graphs and the
  sum-product algorithm,'' \emph{IEEE Trans. on info. theo.}, vol.~47, no.~2,
  pp. 498--519, 2001.

\bibitem{323794}
J.~Shi \emph{et~al.}, ``Good features to track,'' in \emph{Proc. of the {IEEE}
  Int. Conf. on Pattern Recognition}, 1994, pp. 593--600.

\bibitem{LucKan8108}
B.~D. Lucas and T.~Kanade, ``An iterative image registration technique with an
  application to stereo vision,'' in \emph{Proc. of the Intl. Joint Conf. on
  Artificial Intelligence}, Vancouver, Canada, Aug. 1981, pp. 24--28.

\bibitem{6696592}
L.~Heng, B.~Li, and M.~Pollefeys, ``Camodocal: Automatic intrinsic and
  extrinsic calibration of a rig with multiple generic cameras and odometry,''
  in \emph{Proc. of the {IEEE/RSJ} Int. Conf. on Intell. Robots and Syst.},
  2013, pp. 1793--1800.

\bibitem{QinShen17}
T.~Qin and S.~Shen, ``Robust initialization of monocular visual-inertial
  estimation on aerial robots.'' in \emph{Proc. of the {IEEE/RSJ} Int. Conf. on
  Intell. Robots and Syst.}, Vancouver, Canada, 2017.

\bibitem{takasu2009development}
T.~Takasu and A.~Yasuda, ``Development of the low-cost rtk-gps receiver with an
  open source program package rtklib,'' in \emph{Proc. of the Int. Sym. on
  GPS/GNSS}, vol.~1, 2009.

\bibitem{geiger2012we}
A.~Geiger, P.~Lenz, and R.~Urtasun, ``Are we ready for autonomous driving? the
  kitti vision benchmark suite,'' in \emph{Proc. of the {IEEE} Int. Conf. on
  Pattern Recognition}, 2012, pp. 3354--3361.

\bibitem{6906892}
J.~Nikolic, J.~Rehder, M.~Burri, P.~Gohl, S.~Leutenegger, P.~T. Furgale, and
  R.~Siegwart, ``A synchronized visual-inertial sensor system with fpga
  pre-processing for accurate real-time slam,'' in \emph{Proc. of the {IEEE}
  Int. Conf. on Robot. and Autom.}, 2014, pp. 431--437.

\bibitem{blewitt1990automatic}
G.~Blewitt, ``An automatic editing algorithm for gps data,'' \emph{Geophysical
  research letters}, vol.~17, no.~3, pp. 199--202, 1990.

\bibitem{zumberge1997precise}
J.~Zumberge, M.~Heflin, D.~Jefferson, M.~Watkins, and F.~Webb, ``Precise point
  positioning for the efficient and robust analysis of gps data from large
  networks,'' \emph{J. of geophysical research: solid earth}, vol. 102, no.~B3,
  pp. 5005--5017, 1997.

\end{thebibliography}

\end{document}